%% file: main.tex
\documentclass[10pt,twocolumn,letterpaper]{article}

\usepackage{iccv}
\usepackage{times}
\usepackage{epsfig}
\usepackage{graphicx}
\usepackage{amsmath}
\usepackage{amssymb}
\usepackage{iccv}
\usepackage{times}
\usepackage{epsfig}
\usepackage{graphicx}
\usepackage{amsmath}
\usepackage{amssymb}
\usepackage{times}
\usepackage{epsfig}
\usepackage{amsmath}
\usepackage{amssymb}
\usepackage{algorithm, algorithmic}
\usepackage{xcolor}

\usepackage{diagbox}
\usepackage{float}
\usepackage{afterpage}
\usepackage{bm}
\usepackage{times}
\usepackage{epsfig}
\usepackage{graphicx}
\usepackage{subfig}
\usepackage{amsmath}
\usepackage{amssymb}
\usepackage{tabu}
\usepackage{multirow}
\usepackage{color}
\usepackage{tablefootnote}
\usepackage{adjustbox}
\usepackage{bbding}
\usepackage{etoolbox}
\usepackage{booktabs}
\usepackage{soul}
\usepackage{xcolor,colortbl}
\usepackage[pagebackref=true,breaklinks=true,letterpaper=true,colorlinks,bookmarks=false]{hyperref}
\newcommand{\rev}[1]{\textcolor{red}{#1}}
\def\etal{\textit{et al}.}
\def\ie{\textit{i.e.}}
\def\eg{\textit{e.g.}}

\iccvfinalcopy % *** Uncomment this line for the final submission

\def\iccvPaperID{****} % *** Enter the ICCV Paper ID here

\ificcvfinal\fi

\begin{document}

%%%%%%%%% TITLE
\title{\vspace{-5mm}From Continuity to Editability: Inverting GANs with Consecutive Images\vspace{-7mm}}

% \author{Yangyang Xu\\
% South China University of Technology\\
% \and
% Yong Du\\
% Ocean University of China\\
% % First line of institution2 address\\
% {\tt\small csyongdu@ouc.edu.cn}

% \and
% Wenpeng Xiao\\
% South China University of Technology\\
% % First line of institution2 address\\
% {\tt\small 202021044917@mail.scut.edu.cn}

% \and
% Xuemiao Xu\\
% South China University of Technology\\
% % First line of institution2 address\\
% {\tt\small xuemx@scut.edu.cn}

% \and
% Shengfeng He\\
% South China University of Technology\\
% % First line of institution2 address\\
% {\tt\small hesfe@scut.edu.cn}

% }

\author{Yangyang Xu$^1$,
       Yong Du$^2$,
       Wenpeng Xiao$^1$,
       Xuemiao Xu$^{1,3,4,5*}$
       and Shengfeng He$^{1,5}$\thanks{Corresponding authors (\{xuemx,hesfe\}@scut.edu.cn).}
       \\
       \normalsize{$^1$South China University of Technology \hspace{5mm} $^2$Ocean University of China} \\
       \normalsize{$^3$Guangdong Provincial Key Lab of Computational Intelligence and Cyberspace Information}\\
       \normalsize{$^4$State Key Laboratory of Subtropical Building Science}\\
       \normalsize{$^5$Ministry of Education Key Laboratory of Big Data and Intelligent Robot\vspace{-1mm}}
      }

% Xuemiao Xu is with the School of Computer Science and Engineering at South China University of Technology, and also with State Key Laboratory of Subtropical Building Science, Ministry of Education Key Laboratory of Big Data and Intelligent Robot and Guangdong Provincial Key Lab of Computational Intelligence and Cyberspace Information

% \footnote{Corresponding author: (xuemx@scut.edu.cn and hesfe@scut.edu.cn)}
% \author{$^1$School of Computer Science and Engineering, South China University of Technology \\
% $^2$Department of Computer Science and Technology, Ocean University of China}

\setlength{\fboxsep}{0.8pt}
\setlength{\fboxrule}{0pt}

\teaser{\vspace{-3mm}
\centering
%\captionsetup[subfigure]{labelformat=empty}
%\captionsetup[subfloat]{labelformat=empty}
%\rotatebox[origin=c]{90}{Input}
%\rotatebox[origin=c]{90}{Reconstruction}
%\rotatebox[origin=c]{90}{Editing}
\subfloat[Inputs]{\begin{minipage}{.13\linewidth}
\fcolorbox{white}{red}{\includegraphics[width=\textwidth]{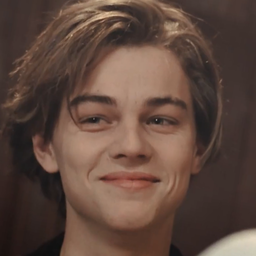}}\\
\fcolorbox{white}{white}{\includegraphics[width=\textwidth]{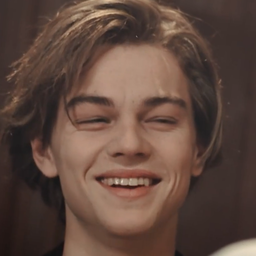}}\\
\fcolorbox{white}{white}{\includegraphics[width=\textwidth]{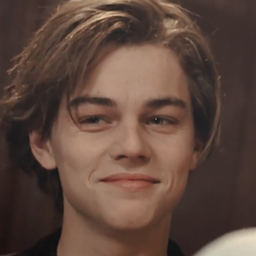}}
\end{minipage}}\hspace{.1mm}
\rotatebox[origin=c]{90}{\hspace{12mm} $\leftarrow$ Gender $\rightarrow$ \hspace{13mm} Reconstruction}
% \rotatebox[origin=c]{90}{\vspace{-10mm} \hspace{-20mm} Editing}
\subfloat[I2S~\cite{abdal2019image2stylegan}]{\begin{minipage}{.13\linewidth}
\centering
\fcolorbox{white}{white}{\includegraphics[width=\textwidth]{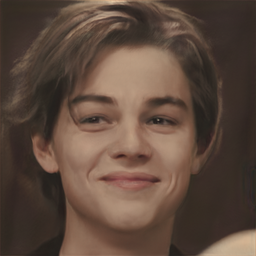}}\\
%\vspace{-1.5mm}MAE: 6.0$\times$e-3\vspace{1.5mm}
\fcolorbox{white}{white}{\includegraphics[width=\textwidth]{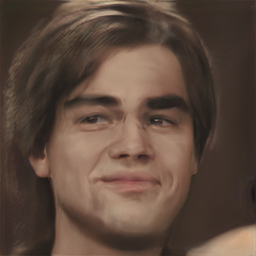}}\\
%\vspace{-1.5mm}NIQE: 10.0\vspace{1.5mm}
\fcolorbox{white}{white}{\includegraphics[width=\textwidth]{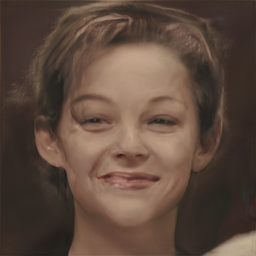}}\\
%\vspace{-1.5mm}NIQE: 10.0\vspace{1.5mm}
\end{minipage}}\hspace{.1mm}
\subfloat[I2S++~\cite{abdal2019image2stylegan2}]{\begin{minipage}{.13\linewidth}
\centering
\fcolorbox{white}{white}{\includegraphics[width=\textwidth]{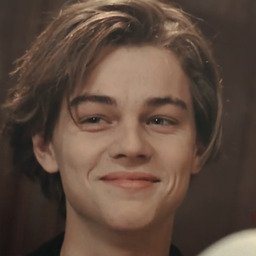}}\\
%\vspace{-1.5mm}MAE: 0.1$\times$e-3\vspace{1.5mm}
\fcolorbox{white}{white}{\includegraphics[width=\textwidth]{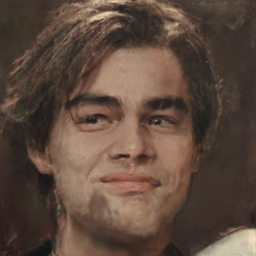}}\\
%\vspace{-1.5mm}NIQE: 10.0\vspace{1.5mm}
\fcolorbox{white}{white}{\includegraphics[width=\textwidth]{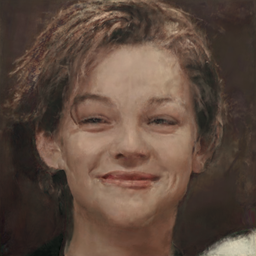}}\\
%\vspace{-1.5mm}NIQE: 10.0\vspace{1.5mm}
\end{minipage}}%\hspace{.1mm}
\subfloat[pSp~\cite{richardson2020encoding}]{\label{fig:teaserc}
\begin{minipage}{.13\linewidth}
\centering
\fcolorbox{white}{white}{\includegraphics[width=\textwidth]{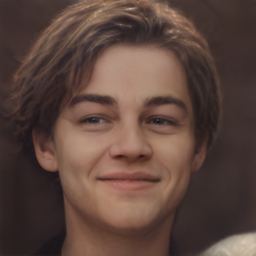}}\\
%\vspace{-1.5mm}MAE: 9.9$\times$e-3\vspace{1.5mm}
\fcolorbox{white}{white}{\includegraphics[width=\textwidth]{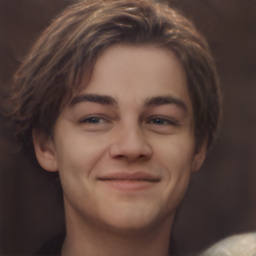}}\\
%\vspace{-1.5mm}NIQE: 10.0\vspace{1.5mm}
\fcolorbox{white}{white}{\includegraphics[width=\textwidth]{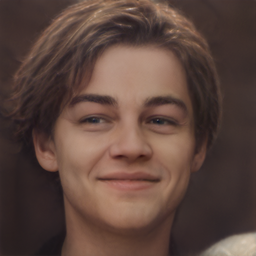}}\\
%\vspace{-1.5mm}NIQE: 10.0\vspace{1.5mm}
\end{minipage}}\hspace{.1mm}
\subfloat[InD~\cite{zhu2020domain}]{\label{fig:teaserd}\begin{minipage}{.13\linewidth}
\centering
\fcolorbox{white}{white}{\includegraphics[width=\textwidth]{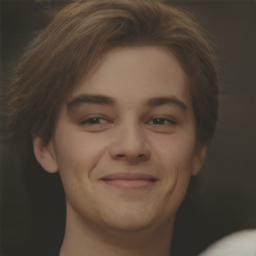}}\\
%\vspace{-1.5mm}MAE: 12$\times$e-3\vspace{1.5mm}
\fcolorbox{white}{white}{\includegraphics[width=\textwidth]{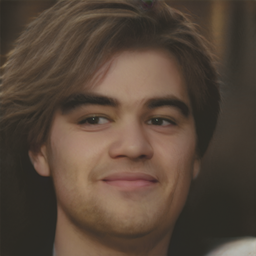}}\\
%\vspace{-1.5mm}NIQE: 10.0\vspace{1.5mm}
\fcolorbox{white}{white}{\includegraphics[width=\textwidth]{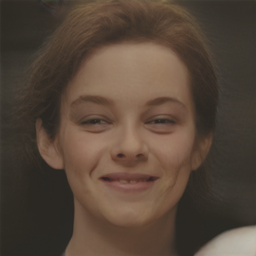}}\\
%\vspace{-1.5mm}NIQE: 10.0\vspace{1.5mm}
\end{minipage}}\hspace{.1mm}
\subfloat[Ours]{\begin{minipage}{.13\linewidth}
\centering
\fcolorbox{white}{white}{\includegraphics[width=\textwidth]{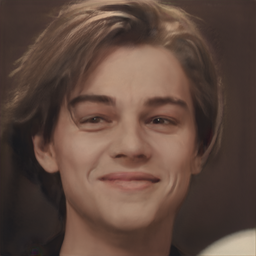}}\\
%\vspace{-1.5mm}\textbf{MAE: 2.5$\times$e-3}\vspace{1.5mm}
\fcolorbox{white}{white}{\includegraphics[width=\textwidth]{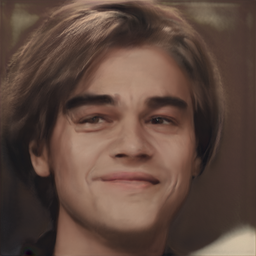}}\\
%\vspace{-1.5mm}\textbf{NIQE: 10.0}\vspace{1.5mm}
\fcolorbox{white}{white}{\includegraphics[width=\textwidth]{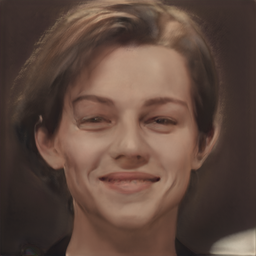}}\\
%\vspace{-1.5mm}\textbf{NIQE: 10.0}\vspace{1.5mm}
\end{minipage}}\vspace{-3mm}
\caption{We propose an alternative GAN inversion method that jointly considers multiple consecutive images. We leverage the inborn continuity between inputs, to simultaneously maximize the reconstruction fidelity (top row) and editability (bottom row) of the inverted latent code. Our method jointly inverts three images from (a), we only show the result of the first one with the red box.}\label{fig:teaser}\vspace{-13mm}
}

\maketitle
\ificcvfinal\thispagestyle{empty}\fi

%%%%%%%%% ABSTRACT
\begin{abstract}\vspace{-4mm}
Existing GAN inversion methods are stuck in a paradox that the inverted codes can either achieve high-fidelity reconstruction, or retain the editing capability. Having only one of them clearly cannot realize real image editing. In this paper, we resolve this paradox by introducing consecutive images (\eg, video frames or the same person with different poses) into the inversion process. The rationale behind our solution is that the continuity of consecutive images leads to inherent editable directions. This inborn property is used for two unique purposes: 1) regularizing the joint inversion process, such that each of the inverted code is semantically accessible from one of the other and fastened in a editable domain; 2) enforcing inter-image coherence, such that the fidelity of each inverted code can be maximized with the complement of other images. Extensive experiments demonstrate that our alternative significantly outperforms state-of-the-art methods in terms of reconstruction fidelity and editability on both the real image dataset and synthesis dataset. Furthermore, our method provides the first support of video-based GAN inversion, and an interesting application of unsupervised semantic transfer from consecutive images. Source code can be found at: \url{https://github.com/cnnlstm/InvertingGANs_with_ConsecutiveImgs}.
% \vspace{-3mm}
%simultaneously maximize the fidelity and editability

%inherent editable direction
\end{abstract}

%%%%%%%%% BODY TEXT
\vspace{-6mm}
\input{1.intro}

\input{2.related}
\input{3.approach}

\input{4.experiments}

\input{5.conclusion}

\small{\textbf{Acknowledgment:} This project is supported by the Key-Area Research and Development Program of Guangdong Province, China (2020B010165004, 2020B010166003, 2018B010107003); National Natural Science Foundation of China (61972162, 61772206, U1611461, 61472145); Guangdong International Science and Technology Cooperation Project (2021A0505030009); Guangdong Natural Science Foundation (2021A1515012625); Guangzhou Basic and Applied Research Project (202102021074); CCF-Tencent Open Research fund (CCF-Tencent RAGR20190112, RAGR20210114); China Postdoctoral Science Foundation (2020M682240, 2021T140631) and Fundamental Research Funds for the Central Universities (202113035).}

{\small
\bibliographystyle{ieee_fullname}
\bibliography{egbib}
}

\newpage
\input{supp3}

\end{document}

%% file: 1.intro.tex
% !TeX root = egpaper_for_review.tex
\section{Introduction}\vspace{-2mm}
Generative Adversarial Networks (GANs)~\cite{NIPS2014_5ca3e9b1,karras2019style,karras2019analyzing} has demonstrated versatile image editing capability, especially by discovering the spontaneously learned interpretable directions that can manipulate corresponding image attributes~\cite{radford2015unsupervised,jahanian2019steerability,shen2019interpreting,goetschalckx2019ganalyze,voynov2020unsupervised}. Concretely, given a random latent code $w$, image editing can be achieved by pushing the latent vector along a specific semantic direction (\eg, age, gender):
\begin{equation}
I{'}=G(w+\alpha \times \mathop{\bold{n}}^{\rightarrow}),
\label{eq1}
\end{equation}
where $I{'}$ is the edited image, $G$ is the generator, $\alpha$ is a scaling factor, and $\bold{\mathop{n}\limits ^{\rightarrow}}$ is the interpretable direction.

As a consequence, recent attempts~\cite{abdal2019image2stylegan,abdal2019image2stylegan2,richardson2020encoding,zhu2020domain} aim to migrate this power to real image editing by inverting an image to the latent code $w$. There are two {keys} for this task, \romannumeral1) whether the inverted code can faithfully reconstruct the original input, and, \romannumeral2) whether the pre-acquired semantic directions can be successfully applied. However, existing methods seem to stuck in a paradox, as achieving one end will inevitably sacrifice the other. As shown in Fig.~\ref{fig:teaser}, I2S, I2S++, and pSp~\cite{abdal2019image2stylegan,abdal2019image2stylegan2,richardson2020encoding} concentrate only on obtaining faithful reconstruction, but the inverted codes show limited editability. In contrast, latent codes obtained from In-domain inversion~\cite{zhu2020domain} (Fig.~\ref{fig:teaserd}) are regularized in the semantically meaningful domain at the expense of fidelity. We argue that balancing these two factors solely based on a single image is extremely challenging, as there is no indicator to shed light on the editable domain in the latent space, preventing the optimization from obtaining a perfect balance between two factors.

In this paper, we resolve the above problem by introducing consecutive images, which can be either a video segment or the same person with different poses, to form a joint optimization process. The rational behind our alternative solution is that the \emph{continuity} brought by consecutive images can be used as an indicator to constrain the \emph{editability}. In particular, to ensure each of the inverted latent codes is editable, we jointly optimize multiple latent codes by enforcing each of them is semantically accessible from one of the other code with a simple linear combination. In addition, we further explore {the sequential continuity} for \emph{fidelity}, by injecting multi-source supervision that common regions of the reconstructed images should be consistent in all the consecutive images. We establish dense correspondences between input images, and then apply the obtained correspondences to warp each of the reconstructions to the neighbors for a consistent and coherence measurement.

To evaluate the proposed method, we construct a real video-based dataset RAVDESS-12, and another consecutive images dataset synthesized by manipulating attributes from the generated images of StyleGAN~\cite{karras2019analyzing}. Extensive experiments demonstrate the superior performance of our solution over existing methods in terms of editability and reconstruction fidelity. Furthermore, our method is capable to perform various applications, \eg, video-based GAN inversion, unsupervised semantic transfer, and image morphing.

In summary, our contributions are three-fold:
\begin{itemize}
    \item We propose an alternative GAN inversion method for consecutive images, and delve deep into the continuity of consecutive images for GAN inversion.
    \item We tailor two novel constraints, one is the mutually accessible constraint that formulates consecutive images inversion as a linear combination process in the latent space to ensure editability, and the inversion consistency constraint that works in the RGB space to guarantee reconstruction fidelity, by measuring reconstruction consistency across {images}.
    \item We demonstrate optimal performances in terms of editability and reconstruction fidelity, and we support various new applications like video-based GAN inversion and unsupervised semantic transfer.
    %\item We present a novel GAN inversion method that combines the latent codes optimization and semantic direction searching together by taking multi-video frames as input.
    %\item Our GAN inversion method performing linear transformation in latent space and transfer dense correspondence in the RGB space, which recovers the more plausible images, and semantic-meaningful latent codes.
    %\item We experiment on various image editing tasks, both qualitative and quantitative results show that our latent codes are more favorable for semantic editing. Besides, our acquired semantic direction can transfer the semantic changes to arbitrarily images.
\end{itemize}

%% file: 2.related.tex
% !TeX root = egpaper_for_review.tex
\vspace{-2mm}\section{{Related Work}}\vspace{-2mm}

\textbf{Image Editing via Latent Space Exploration.}
Generative models show great potential in synthesizing versatile images by taking random latent codes as inputs. Recent works show that the latent space of a pre-trained GAN encodes rich semantic directions. Varying the latent codes with a specific direction can edit the image with the target attribute. In particular, Radford~\etal~\cite{radford2015unsupervised} observe that there are directions in the latent space corresponding to adding smiles or glasses on the faces. Ganalyze~\etal~\cite{goetschalckx2019ganalyze} explore the memorability direction in the latent space by a fixed assessor. Jahanian~\etal~\cite{jahanian2019steerability} study the steerability of GANs to fit some image transformations. Shen~\etal~\cite{shen2019interpreting} explore the semantic boundary in the latent space of the binary attributes. Voynov~\etal~\cite{voynov2020unsupervised} discover the semantic directions hidden in the latent space by an unsupervised model-agnostic procedure. Varying the latent codes with such directions can manipulate the corresponding attributes of the output images. It is natural to transferring these techniques on real image editing. Before that, it is required to invert a real image back to the latent code.

\textbf{GAN Inversion.}
To realize real image editing, GAN inversion methods are proposed to inference a latent code of an input image based on the pre-trained GAN~\cite{perarnau2016invertible,zhu2016generative,abdal2019image2stylegan,creswell2018inverting,pan2020exploiting}. These methods can be categorized into two classes, optimization-based and encoder-based. The former individually optimizes the latent code for a specific image, with a concentration on pixel-wise reconstruction~\cite{abdal2019image2stylegan,abdal2019image2stylegan2,creswell2018inverting,NEURIPS2018_e0ae4561}. However, ensuring reconstruction fidelity cannot guarantee the output latent code is editable by the learned directions. On the other hand, encoder-based methods train a general encoder that maps real images to latent codes~\cite{zhu2020domain,richardson2020encoding}. Especially, In-Domain GAN inversion~\cite{zhu2020domain} combines the learned encoder with an optimization process to align the encoder with the semantic knowledge of the generator. However, existing methods do not resolve the problem of editable domain in the latent space, thus they cannot achieve a perfect balance between editability and fidelity. We aim to solve this problem from a novel view of considering multiple consecutive images.

%% file: 3.approach.tex
\vspace{-2mm}\section{Method}\vspace{-2mm}

\begin{figure*}[t]
\centering
\includegraphics[width=0.99\linewidth]{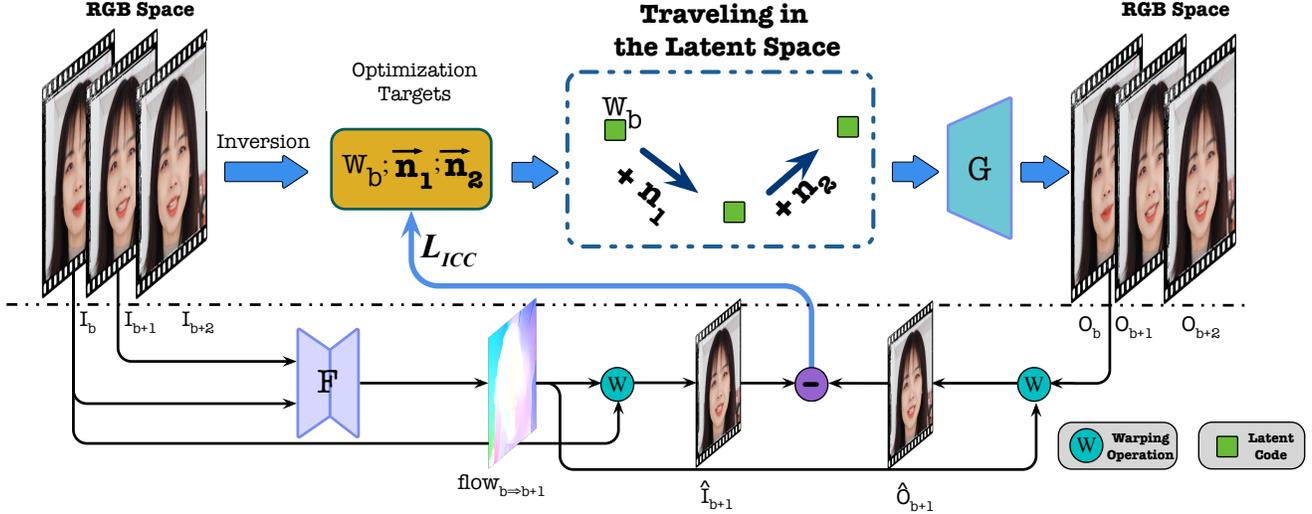}
\vspace{-3mm}
\caption{The pipeline of the proposed consecutive images based GAN inversion. $G$ is a pretrained generator of StyleGAN and $F$ is a pretrained FlowNet for calculating the bidirectional optical flow. The upper part shows the Mutually Accessible Constraint. Given consecutive images as input, we enforce each of them is semantically accessible from one of the other codes with a simple linear combination. Both~$w_{1}$ and~$\bold{\mathop{n}\limits ^{\rightarrow}}_{k}$ are the optimization targets. The bottom shows the Inversion Consistency Constraint in the RGB space. Note here we only show the forward flow among $I_b$ and $I_{b+1}$ when calculating the $\mathcal{L}_{ICC}$ for simplicity, meanwhile, $\mathcal{L}_{C}$ and $\mathcal{L}_{P}$ are also omitted.}
\label{fig:overview}\vspace{-5mm}
\end{figure*}

\subsection{Overview}\vspace{-2mm}
\label{sec:igi}
The editability of the latent code and the fidelity of the reconstruction are the two vital factors that affect the performance of GAN inversion. To satisfy both sides, we exploit the continuity brought by consecutive images which depict a same subject with different variations. The pipeline of our approach is illustrated in Fig.~\ref{fig:overview}. Given a sequence of consecutive images as input, the proposed method aims to seek their optimal latent codes in the latent space, and then feed them into a pretrained and fixed generator for reconstruction. In particular, \romannumeral1) we define a linear combination mechanism among consecutive images, which would facilitate the editability of the latent codes via a joint optimization with semantic directions, and \romannumeral2) we establish a consistency constraint in the RGB space between the warped results of the reconstructed images and their corresponding originals, promoting the fidelity of the reconstruction. Note that we choose the generator of StyleGAN~\cite{karras2019style} as the pretrained one in our model.

\subsection{Consecutive Images Based GAN Inversion}\label{sec.method}\vspace{-2mm}
\textbf{Mutually Accessible Constraint.}
For each image in an input set, it may gradually change into the others, just as shown in Fig.~\ref{fig:overview}, the mouth opens progressively. Or it can vary to the others in a drastic way, such as the same person at very different poses. In either case, given the first image $I_1$, it can be intuitively assumed that the latent codes of the other images are linear combinations of that of the starting image along with a specific semantic direction $\bold{\mathop{n}\limits^{\rightarrow}}$ (\eg, expression, pose). Then, the latent code $w_{1+k}$ of the image $I_{1+k}$ can be formulated as follows:
\begin{equation}
w_{1+k} = w_{1}+\alpha_k \times \mathop{\bold{n}}^{\rightarrow}, \quad k=1,2,\cdots,T-1,
\label{eq4}
\end{equation}
where $T$ denotes the total number of images. However, the assumption is too strong since consecutive images are likely to vary from one to the others in different attributes. On the other hand, the specific semantic direction $\bold{\mathop{n}\limits^{\rightarrow}}$ can be predefined~\cite{shen2019interpreting,voynov2020unsupervised,shen2021closedform} , but the semantically equivalent input images are required. To cope with images with arbitrary semantic variations, we regard the direction as one of our optimization targets. Note that the scaling factors $\alpha_k$-s are also learnable, and therefore, we absorb them into the direction and reformulate Eq.(\ref{eq4}) as the proposed mutually accessible constraint, which is as follows:
\begin{equation}
% w_{1+k} = w_{1}+\mathop{\bold{n}_k}^{\rightarrow},
w_{1+K} = w_{1}+ \sum_{k=1}^K \mathop{\bold{n}_k}^{\rightarrow}.
\label{eqxxx}
\end{equation}

% $\alpha_k$\text{'s}
In this way, we can figure out the latent codes of all the images via a joint optimization of $w_1$ and $\bold{\mathop{n}\limits^{\rightarrow}}_k$-s. Such a simple linear combination mechanism can promote the editability of the latent codes. The main reasons are that, \romannumeral1) each of the inverted latent codes can be regarded as an edited code with respect to one of the others, and \romannumeral2) if the images vary in a specific semantic direction, the scales of variations could be learned adaptively, and more importantly, \romannumeral3) it is able to deal with the variations of different attributes among consecutive images. Moreover, the learned $\bold{\mathop{n}\limits^{\rightarrow}}_k$-s have the potential to be transferred to other latent codes as predefined semantic directions.

\textbf{Inversion Consistency Constraint.}
Once we have the latent codes, we feed them into the generator $G(\cdot)$ to reconstruct consecutive images. For a certain base image $I_b$, its reconstruction $O_b$ with regard to the latent code $w_b$ is calculated by
\begin{equation}
O_b = G(w_b), \quad b=1,2,\cdots,T.
\label{eqxxxx}
\end{equation}

In order to ensure a fidelity of the reconstructions, we particularly consider an inversion consistency between common regions of the reconstructed images and the input consecutive images. Specifically, we tailor an inversion consistency constraint loss based on bidirectional flows in the RGB space. As shown in the bottom of Fig.~\ref{fig:overview}, first, the forward flows $f_{b \Rightarrow b+k}$ between base image $I_{b}$ and the other images $I_{b+k}$ can be calculated by a pretrained FlowNet2~\cite{ilg2017flownet} $F(\cdot)$, which is formulated as follows:
\begin{equation}
f_{b \Rightarrow b+k} = F(I_b,I_{b+k}),
\label{eq6}
\end{equation}
where $k\in\mathbb{Z}\cap[1-b,T-b]-\{0\}$. Then we warp $I_{b}$ with this flow to form the warped images $\hat{I}_{b+k}$, which can be presented as follows:
\begin{equation}
\hat{I}_{b+k} = warp(I_b,f_{b \Rightarrow b+k}).
\label{eq7}
\end{equation}
Also, we have the warped results $\hat{O}_{b+k}$ of the recovered images $O_b$, which is presented as follows:
\begin{equation}
\hat{O}_{b+k} = warp(O_b,f_{b \Rightarrow b+k}).
\label{eq8}
\end{equation}
Since consecutive images describe a same subject, an inherent relationship should be existed among the generated warpings $\{\hat{I}_{b+k}\}$ for each base image $I_b$. And this relationship should be transferred to the other warpings $\{\hat{O}_{b+k}\}$. In the same way, we can calculate the backward flow $f_{b+k \Rightarrow b}$ between $I_{b+k}$ and $I_{b}$, and the corresponding warpings $\{\hat{I}_b^{b+k}\}$, $\{\hat{O}_b^{b+k}\}$. By iterating over all the values of $b$ and $k$, we inject multi-source supervision from the warpings of input images to confine the reconstructions, and the proposed inversion consistency constraint loss ${\cal L}_{ICC}$ is given by
\begin{equation}
\mathcal{L}_{ICC} = \sum_{b}\sum_{k}(\|\hat{I}_{b}^{b+k} - \hat{O}_{b}^{b+k} \|_{2}+\|\hat{I}_{b+k}-\hat{O}_{b+k}\|_{2}),
\label{eq9}
\end{equation}
where $\|\cdot\|_{2}$ denotes a pixel-wise $L_2$ distance.

Moreover, we consider to maintain a pixel-wise consistency between the input images and its corresponding recoveries. A pixel-wise consistency loss ${\cal L}_C$ is therefore introduced in our objective, that is
\begin{equation}
\mathcal{L}_{C} = \sum_{m\in q\cup\{0\}} \|O_{1+m}-I_{1+m}\|_{2},
\label{eq10}
\end{equation}
To guarantee a fine visual perception of the recontructions, we also utilize a perceptual loss ${\cal L}_P$ which is formulated as follows:
\begin{equation}
\begin{aligned}
\mathcal{L}_{P}  = \frac{1}{4} \sum_{j=1}^4 \sum_{m\in q\cup\{0\}} \|\Phi_{V}^j(I_{1+m})-\Phi_{V}^j(O_{1+m}) \|_{2},
\label{eq11}
\end{aligned}
\end{equation}
where $\Phi_{V}^j(\cdot)$ denotes the $j$th layer of a pretrained VGG-16 network, and we follow Abdal \etal~\cite{abdal2019image2stylegan} to select the features produced by the \texttt{conv1\_1}, \texttt{conv1\_2}, \texttt{conv3\_2} and \texttt{conv4\_2} layers of VGG-16 for modeling the loss.

Finally, the whole objective function $\cal L$ is defined as follows:
\begin{equation}
{\cal L}=\lambda_1{\cal L}_{ICC}+\lambda_2{\cal L}_{C}+\lambda_3{\cal L}_{P},
\end{equation}
where $\lambda$s denote the balance factors. Then the latent code $w_{1}$ and the directions $\bold{\mathop{n}\limits ^{\rightarrow}}_k$ can be optimized by
\begin{equation}
\label{eq12}
\{w^*_{1}, \bold{\mathop{n}\limits ^{\rightarrow}}_k^*\}= \mathop{\arg\min}_{\{w_{1},\bold{\mathop{n}\limits ^{\rightarrow}}_k\}} \lambda_{1} \mathcal{L}_{ICC}+ \lambda_{2} \mathcal{L}_{C}+ \lambda_{3} \mathcal{L}_{P}.%+ \lambda_{4} \mathcal{L}_{pp},
\end{equation}

Note that we follow \cite{abdal2019image2stylegan} that initialize $w_{1}$ with the mean latent vector of $\mathcal{W+}$ space. And the direction $\bold{\mathop{n}\limits^{\rightarrow}}_k$-s are initially set to zero and updated during optimization.

%% file: 4.experiments.tex
% !TeX root = egpaper_for_review.tex
\vspace{-2mm}\section{Experiments}\vspace{-2mm}

\begin{table*}[t]
   \caption{Comparisons with existing GAN inversion methods on image reconstruction with four metrics on two datasets. $\downarrow$ denotes the lower the better and the best results are marked in \textbf{bold}.}
       \vspace{-0.6cm}
       \begin{center}
       \setlength{\tabcolsep}{0.2cm}{
       \begin{tabular}{c|c|c|c|c|c|c|c|c}
           \hline
           \multirow{1}{*}{\diagbox{Methods}{Metrics}}       & \multicolumn{4}{c|}{RAVDESS-12 Dataset} & \multicolumn{4}{c}{Synthesized Dataset} \\
           \cline{2-9}
          &{NIQE$\downarrow$}   &{FID$\downarrow$}  &{LPIPS$\downarrow$}   &{MSE$\downarrow$($\times$e-3)}  &{NIQE$\downarrow$}  &{FID$\downarrow$}  &{LPIPS$\downarrow$}  &{MSE$\downarrow$($\times$e-3)}   \\
           \hline
           I2S~\cite{abdal2019image2stylegan}   &3.770    &16.284  &0.162   &8.791                      &3.374&48.909&0.271&35.011         \\
           pSp~\cite{richardson2020encoding}    &3.668    &29.701  &0.202   &22.337                     &3.910&84.355&0.391&46.244                \\
           InD~\cite{zhu2020domain}             &3.765    &18.135  &0.193   &9.963                      &3.152&42.773&0.352&44.645                  \\
           Ours    &\textbf{3.596}     &\textbf{13.136}   &\textbf{0.148}  &\textbf{5.972}             &\textbf{2.807}&\textbf{37.225}&\textbf{0.250}&\textbf{24.395}    \\
           \hline
           \hline
           % I2S++~\cite{abdal2019image2stylegan2} &3.358    &0.320   &\textbf{0.003}   &0.174     &2.644         &2.967         &\texfbf{0.014} &1.458             \\
           I2S++~\cite{abdal2019image2stylegan2} &3.358    &0.320   &\textbf{0.003}   &0.174     &2.644         &2.967         &\textbf{0.014} &1.458             \\

           Ours++    &\textbf{3.352}     &\textbf{0.311}    &\textbf{0.003}  &\textbf{0.165}     &\textbf{2.597}&\textbf{2.897}&\textbf{0.014} &\textbf{1.432}    \\

           \hline
       \end{tabular}
       }
      \end{center}
      \label{table:reconstruction}
      \vspace{-8mm}
\end{table*}

\subsection{Implementation Details}\vspace{-2mm}

We implement the proposed method in Pytorch on a PC with an Nvidia GeForce RTX 3090. We utilize the generator of StyleGAN~\cite{karras2019style} pre-trained on the FFHQ dataset~\cite{karras2019style} with the resolution of $1024\times 1024$. The latent codes and semantic directions are optimized using Adam optimizer~\cite{kingma2014adam}. We follow~\cite{abdal2019image2stylegan} that use 5000 gradient descent steps with a learning rate of 0.01, ${\beta}_1 = 0.9$, ${\beta}_2 = 0.999$, and ${\epsilon} = 1e^{-8}$. We empirically set the balancing weights in Eq.~\eqref{eq12} as $\lambda_{1}=1$, $\lambda_{2}=1$ and $\lambda_{3}=1$. And we set $T=5$ in Eq.~\eqref{eqxxxx}, which indicates 5 consecutive images are contained in each input sequence. %\rev{Besides, we set the value scale of $\alpha$ in Eq.~\eqref{eq4} as [-1,1], and vary it uniformly to forming the continuous frames during the optimization.}

\subsection{Experimental Settings}\vspace{-2mm}

\textbf{Datasets. }
We first conduct our experiments on the RAVDESS dataset~\cite{livingstone2018ryerson} with real videos. The original RAVDESS dataset contains 2,452 videos with 24 subjects speaking and singing with various semantic expressions. We select 12 videos of them for evaluation, resulting in 1,454 frames, we name this dataset as RAVDESS-12 Dataset. %All frames are aligned before feeding to the method.
Since there are no ground truth latent codes for real images, we cannot subjectively evaluate the inverted code and its editability in the latent space. On the other hand, it is demonstrated that the learned semantic directions work very well in generated images. As a result, we construct a synthesized dataset, containing 1000 samples that were randomly generated by StyleGAN. For each sample, we vary its latent code with 5 random combinations of the $\alpha$ value (ranging from -3 to 3) and semantic direction (acquired from InterfaceGAN~\cite{shen2019interpreting}), producing 5000 images. We record the latent codes of the original samples, the corresponding editing specifications, and the edited latent codes for evaluating the editability. GAN inversion methods will invert the original samples and edit them to target attributes for comparisons.

\textbf{Competitors.}
We mainly compare with four GAN inversion methods: Image2StyleGAN (I2S)~\cite{abdal2019image2stylegan}, Image2StyleGAN++ (I2S++)~\cite{abdal2019image2stylegan2}, In-domain Inversion (InD)~\cite{zhu2020domain}, and pSp network~\cite{richardson2020encoding}. All the methods are inverted to the same $\mathcal{W+}$ latent space of StyleGAN, applying the same directions for editing. It is worth noting that I2S++ introduces the additional noise space $\mathcal{N}$ for small details recovery. A main problem is that, the inverted two latent codes in the $\mathcal{W+}$ and $\mathcal{N}$ spaces are highly coupled, but the learned semantic directions are optimized in $\mathcal{W+}$ only. Applying them changes the $\mathcal{W+}$ space latent code but leaves the noise vector unchanged, these unpaired vectors yield ``ghosting'' artifacts after editing (see Fig.~\ref{fig:semantic_exp}). As a result, we mainly use it for reconstruction comparisons, and we also extend our method to include the noise space, named as Ours++.

\input{inversion_quality}

%In particular, the pSp~\cite{richardson2020encoding} learns an encoder that maps the image to $\mathcal{W+}$ latent space. InD~\cite{zhu2020domain} encodes the input image to an initial latent code $w$ and further optimize it. Image2StyleGAN~\cite{abdal2019image2stylegan} optimizes the latent code $w$ in the disentangled $\mathcal{W+}$ space of StyleGAN. After obtaining the $w$ code, I2S++ further optimizing the noise latent code $n$ in the $\mathcal{N}$ space for recovering the high-frequency information of the image. Following this work, we also optimize $n$ based on our inverted~$w$ code (denoted by Ours++). In this way, the inverted $w$ and $n$ of the same image are highly-related. However, all the semantic editing tasks are conducted in the $\mathcal{W+}$ space, when the $w$ code is transformed, the high-frequency information encodes in the $n$ will yield the artifacts (See the ``ghost'' in i2s++ and Ours++ of Fig.~\ref{table:semantic}), thus, we only compare I2S++ in the image reconstruction task. The detailed information of competitors are shown in Tab.~\ref{table:competitors}.

%Besides, the original in-domain inversion~\cite{zhu2020domain} is experiment with the resolution of $256\times 256$, and pSp network is working on the generator of StyleGAN2~\cite{karras2019analyzing}. We re-train their models based on their official codes to have the same experimental setting as ours.

%\input{compare_table}

\textbf{Evaluation Metrics. }
For the quantitative comparisons, we use four metrics, Naturalness Image Quality Evaluator (NIQE)~\cite{mittal2012making}, Fréchet Inception Distance (FID)~\cite{heusel2017gans}, Learned Perceptual Image Patch Similarity (LPIPS)~\cite{zhang2018unreasonable}, and pixel-wise Mean Square Error (MSE), for evaluating the reconstruction fidelity. Especially, FID computes the Wasserstein-2 distance between the distribution of input and output images. NIQE evaluates the quality perceived by a human, which is a completely blind assessment with no request for the GT image. Since there is no GT for the semantic editing task on the real RAVDESS-12 Dataset, we use FID and NIQE to evaluate real image editing results.
% {}

% \subsection{Distance Analysis Between $w$ And $n$}
% As proved in InterfaceGAN~\cite{shen2019interpreting}, for a latent code $z$ sampled from Gaussian noise space~$\mathcal{N}$, and a semantic boundary~$\bold{n}$~which satisfies with $\bold{n^Tn}=1$, when $z$ is away from boundary~$\bold{n}$, then~$z$~is hard to be edited by this boundary. Did this property could be extended to the disentangled~$\mathcal{W+}$ space?

\subsection{Evaluation on Image Reconstruction}\vspace{-2mm}
\textbf{Quantitative Evaluation. }We first evaluate the reconstruction fidelity of the inverted codes. Quantitative comparisons are shown in Tab.~\ref{table:reconstruction}. We can see that our method significantly outperforms three editable GAN inversion methods (upper part) on both the real dataset and the synthesized one. Especially for the pixel-wise difference metric MSE, we largely improve the state-of-the-art by 31\%. This indicates that the proposed joint optimization successfully incorporates complementary information from neighboring images. Besides, by involving the noise space $\mathcal{N}$, I2S++ and Ours++ achieve the most faithful reconstruction compared with other methods. Thanks to the inter-image coherence, we further push the reconstruction record a bit.

%Except the numerical analysis of the recovered images, we also analyze the semantic of inverted codes. As described in InterfaceGAN~\cite{shen2019interpreting}, for a binary attribute, there is a hyperplane in the latent space and the latent codes of the same side share the same attribute. We use this hyperplane to evaluate the semantic of the inverted codes.

%In particular, we conduct this experiment on RAVDESS-12 Dataset, we first use a pre-trained binary classifiers to predict the attributes of the real images, and take the predicted results as the ground truth. We chose gender(male,\vs,female), smile(absence \vs presence) and age(young \vs old) for analyzing. Then we use InterfaceGAN~\cite{shen2019interpreting} for searching the semantic boundaries and use it evaluate the attributes of inverted code. The receiver operating characteristic (ROC) curves are shown in Fig.~\ref{fig:semantic_roc}. We can seen that our method achieves the best performance under the evaluation of boundaries, \rev{which indicates that our inverted codes inherit more semantic knowledges from the input images, which improves the editing capability.}

%\input{roc}

\textbf{Qualitative Evaluation.} Qualitative comparisons are shown in Fig.~\ref{fig:reconstruction_exp}. We can see that the Image2StyleGAN cannot recover the image color correctly. Meanwhile, the pSp and InD cannot recover the finest facial details of the original images (see teeth in the first row). Compared with the above three methods, our method can reconstruct faithful appearance details. Unsurprisingly, I2S++ recovers the finest details among all competitors. That mainly because of their optimization performed in the $\mathcal{N}$ space encodes high-frequency information. We also depict our results optimized in $~\mathcal{N}$ space, and we preserve the original color better than I2S++~(see the second row).

\input{semantic_editing1}

\begin{table}[t]
   \caption{Quantitative evaluation on real image manipulation with two blind metrics on the RAVDESS-12 Dataset. $\downarrow$ denotes the lower the better and the best results are marked in \textbf{bold}.}
       \vspace{-0.6cm}
       \begin{center}
       \setlength{\tabcolsep}{0.1cm}{
       \begin{tabular}{c|c|c|c|c}
           \hline
           {Metrics}        &{I2S~\cite{abdal2019image2stylegan}}      &{pSp~\cite{richardson2020encoding}}   &{InD~\cite{zhu2020domain}}   & Ours\\
           \hline
           NIQE  $\downarrow$         &   3.776         &5.242          &3.693                &\textbf{3.254}          \\
           FID  $\downarrow$          &   21.609        &30.128         &19.271               &\textbf{15.482}         \\
           \hline
       \end{tabular}
       }
      \end{center}
      \label{table:semantic}
      \vspace{-8mm}
\end{table}

\input{semantic_editing2}

\subsection{Evaluation on Image Editing}\vspace{-2mm}

In this section, we evaluate our GAN inversion method on real image editing as well as synthesized images. We conduct two editing tasks based on the inverted latent codes, the first one is semantic manipulation and the second is image morphing.

\subsubsection{Semantic Manipulation}\vspace{-2mm}

Semantic manipulation aims to edit a real image by varying its inverted codes along with a specific semantic direction. We use five semantic directions (\ie, gender, pose, smile, eyeglasses, and age) acquired by~\cite{shen2019interpreting} in the experiment.% and we set $\alpha=1$ in Eq.~\eqref{eq1} for positive and $\alpha=-1$ for negative.

\begin{table}[t]
   \caption{Quantitative evaluation on image manipulation with four metrics on the Synthesized Dataset. $\downarrow$ denotes the lower the better and the best results are marked in \textbf{bold}.}
       \vspace{-0.6cm}
       \begin{center}
       \setlength{\tabcolsep}{0.1cm}{
       \begin{tabular}{c|c|c|c|c}
           \hline
           {Metrics}        &{I2S~\cite{abdal2019image2stylegan}}      &{pSp~\cite{richardson2020encoding}}   &{InD~\cite{zhu2020domain}}   & Ours\\
           \hline
           NIQE  $\downarrow$                      &   3.390         &3.917          &3.193      &\textbf{3.163}          \\
           FID  $\downarrow$                       &   35.894        &58.342         &48.867     &\textbf{33.872}         \\
           LPIPS  $\downarrow$                     &   0.399         &0.452          &0.424      &\textbf{0.347}         \\
           MSE$\downarrow$($\times$e-3)          &   89.671        &126.642           &101.563      &\textbf{70.224}         \\
           \hline
       \end{tabular}
       }
      \end{center}
      \label{table:semantic_g}
      \vspace{-8mm}
\end{table}

\textbf{Qualitative Evaluation.} The qualitative comparisons on real data are shown in Fig.~\ref{fig:semantic_exp}. We can see that our manipulated faces have visually more plausible results than those of the competitors. In particular, the manipulated results gained by Image2StyleGAN~\cite{abdal2019image2stylegan} present noisy artifacts with pose changes and glasses are entangled with the age attributes, revealing that the edited latent codes are escaped from the editable domain. Similar situations can be found in I2S++~\cite{abdal2019image2stylegan2}. The manipulated faces based on pSp~\cite{richardson2020encoding} are almost unchanged. This is because pSp focuses on learning a direct mapping from the input to latent code, ignoring the editability. This problem is addressed by In-domain inversion~\cite{zhu2020domain}, but it also sacrifices the reconstruction quality. In contrast, thanks to the jointly considered inherent editability constraint between consecutive images, our inverted latent codes are more semantically editable, leading to more disentangled manipulations. On the other hand, the noise space optimization methods (right part of Fig.~\ref{fig:semantic_exp}) show obvious noise artifacts than the others with pose changes, this is because the pre-optimized noise vector is not suitable for the edited latent vector $w$. However, Ours++ can disentangle gender from glasses better than the inverted vector from I2S++.

To evaluate whether the obtained inversion can be edited by arbitrary directions, we force the editing direction different from the semantic changes contained the input sequence on the synthesized dataset. The qualitative comparisons on the synthesized data are shown in Fig.~\ref{fig:semantic_exp_g}. Similar to the evaluation on real data, I2S produces obvious artifacts, pSp fails to edit the results, and InD cannot preserve the original identity. Our manipulated results are more similar with the GTs, which indicates that our inverted codes are much closer with the GT latent codes and also inherit their editability.

\textbf{Quantitative Evaluation.} We present the quantitative comparisons in Tab.~\ref{table:semantic} and Tab.~\ref{table:semantic_g}. Our method achieves the best results on both the RAVDESS-12 Dataset and the Synthesized Dataset. In particular, for the blind metric NIQE, our edited results achieve 13.8\% improvement over the state-of-the-art method, which indicates that our editing is more visually plausible. Quantitative results on the Synthesized Dataset can evaluate whether the inverted codes are close enough with the GT code such that we can reuse their semantic information. From two non-blind metrics LPIPS and MSE, we can see that our edited results are very similar to the GTs. Thanks to our semantically accessible regularization in the latent space, our inverted latent codes show a strong editability compared with the competitors.

\input{image_morphing}

\subsubsection{Image Morphing}\vspace{-2mm}

Image morphing aims to fuse two images semantically by interpolating their latent codes. It is another way to evaluate whether the inverted codes indeed lie in the latent space and reuse the semantic knowledge. For the high-quality inverted codes, their interpolated results should also stay in the editable domain and the semantic varies continuously. Qualitative comparisons are shown in Fig.~\ref{fig:image_morphing}. We can see that the morphing results produced by Image2StyleGAN~\cite{abdal2019image2stylegan} have noticeable artifacts. Meanwhile, the results produced by pSp~\cite{richardson2020encoding} are unrealistic with the unnatural hairs. In contrast, our method presents high-quality results with a continuous morphing process. We also present the quantitative evaluation on the morphing task in Tab.~\ref{table:morphing} and Tab.~\ref{table:morphing_g}, we can see that our inversion results outperform the other inversion methods on both the real dataset and the synthesized one.

\begin{table}[t]
   \caption{Quantitative evaluation on image morphing with two blind metrics on the RAVDESS-12 Dataset. $\downarrow$ denotes the lower the better and the best results are marked in \textbf{bold}.}
       \vspace{-0.6cm}
       \begin{center}
       \setlength{\tabcolsep}{0.1cm}{
       \begin{tabular}{c|c|c|c|c}
           \hline
           {Metrics}        &{I2S~\cite{abdal2019image2stylegan}}      &{pSp~\cite{richardson2020encoding}}   &{InD~\cite{zhu2020domain}}   & Ours\\

           \hline
           NIQE  $\downarrow$   &4.255        &5.350       &4.051    &\textbf{3.688}                    \\
           FID  $\downarrow$    &40.627        &38.474     &38.925           &\textbf{37.695}         \\

           \hline
       \end{tabular}
       }
      \end{center}
      \label{table:morphing}
      \vspace{-7mm}
\end{table}

\begin{table}[t]
   \caption{{Quantitative evaluation on image morphing with four metrics on the Synthesized Dataset. $\downarrow$ denotes the lower the better and the best results are marked in \textbf{bold}}.}
       \vspace{-0.6cm}
       \begin{center}
       \setlength{\tabcolsep}{0.1cm}{
       \begin{tabular}{c|c|c|c|c}
           \hline
           {Metrics}        &{I2S~\cite{abdal2019image2stylegan}}      &{pSp~\cite{richardson2020encoding}}   &{InD~\cite{zhu2020domain}}   & Ours\\

           \hline
           NIQE  $\downarrow$   &3.389            &3.800            &3.212                 &\textbf{3.115}                    \\
           FID  $\downarrow$    &31.776        &30.192                          &{21.901}     &\textbf{18.621}    \\
           LPIPS  $\downarrow$  &0.472         &0.467            &{0.469}          &\textbf{0.402}         \\
           MSE$\downarrow$($\times$e-3)   &141.432         &121.834            &{125.674}          &\textbf{98.354}         \\

           \hline
       \end{tabular}
       }
      \end{center}
      \label{table:morphing_g}
      \vspace{-8mm}
\end{table}

\input{semantic_transfer}

\subsection{Semantic Transfer}\vspace{-2mm}

As discussed in Sec.~\ref{sec.method}, both the latent code~$w$~and the semantic direction~$\bold{\mathop{n}\limits ^{\rightarrow}}$ can be unsupervisedly obtained after inversion. Besides the latent code, our acquired direction~$\bold{\mathop{n}\limits ^{\rightarrow}}$~represents the semantic changes of the input images. Given the input images as reference, we can transfer its semantic changes to the target faces. %In particular, given the latent code of a target image and a semantic direction obtained from the reference image set, we perform semantic manipulation following Eq.~\ref{eq1}.

The transfer results are shown in Fig.~\ref{fig:semantic_trans}. We can see that the semantic attributes of target faces are modified following the reference image set. Note that there are more than one attribute has been changed in the reference. For example, in the right example, the mouth and pose varies simultaneously but we can still capture those changes. This shows that our acquired direction is disentangled with the referenced and can be applied on other faces. Other than existing supervised~\cite{shen2019interpreting,shen2020interfacegan} or unsupervised~\cite{shen2021closedform,voynov2020unsupervised} learning of interpretable directions, this sheds light on a new exemplar-based learning of semantic direction.

% That shows our acquired direction can be applied on other faces, compared with related works that searching the semantic directions, our method is much easier

% That shows our acquired direction can be applied on other faces, which directs another way for searching the semantic directions beyond the recent works~\cite{shen2019interpreting,shen2021closedform,voynov2020unsupervised}.

% which can be used used for transfer the semantics on target faces

\subsection{Ablation Study}\vspace{-2mm}

In this section, we analyze the efficacy of our two components: mutually accessible constraint (MAC) and inversion consistency constraint (ICC). Note that without these two components, our method equals to the Image2StyleGAN inversion and we set it as our baseline. By unplugging one of these two constraints, we yield two variants of ``$w/o$ MAC'' and ``$w/o$ ICC''. In this case, $\mathcal{L}_{ICC}$ is removed and all the latent codes are optimized simultaneously.

\begin{table}[t]
   \caption{Ablation study on image reconstruction with four metrics. $\downarrow$ denotes the lower the better and the best results are marked in \textbf{bold}.}
       \vspace{-0.6cm}
       \begin{center}
       \setlength{\tabcolsep}{0.05cm}{
       \begin{tabular}{c|c|c|c|c}
           \hline
           {Variants}    &{NIQE$\downarrow$}   &{FID$\downarrow$} &{LPIPS$\downarrow$}   &MSE$\downarrow$($\times$e-3) \\

           \hline
           Baseline                         &3.770              &16.284            &0.162                     &8.791                                \\
           $w/o$ MAC     &3.685               &13.375            &0.151                       &8.065                             \\
           $w/o$ ICC      &3.765                 &14.791            &0.160                       &8.508                               \\
           Ours            &\textbf{3.596}         &\textbf{13.136}   &\textbf{0.148}            &\textbf{5.972}       \\
           \hline
       \end{tabular}
       }
      \end{center}
      \label{table:ablation_inversion}
      \vspace{-7mm}
\end{table}

\begin{table}[t]
   \caption{Ablation study on semantic manipulation with two blind metrics. $\downarrow$ denotes the lower the better and the best results are marked in \textbf{bold}.}
       \vspace{-0.6cm}
       \begin{center}
       \setlength{\tabcolsep}{0.1cm}{
       \begin{tabular}{c|c|c|c|c}
           \hline
           {Metrics}    &Baseline   &$w/o$ MAC    &$w/o$ ICC   &Ours     \\
           \hline
           {NIQE$\downarrow$}               &3.776    &3.659     &3.398    &\textbf{3.254}\\
           {FID$\downarrow$}                &21.609   &16.121.   &17.274   &\textbf{15.482}\\
           \hline
       \end{tabular}
       }
      \end{center}
      \label{table:ablation_semantic}
      \vspace{-9mm}
\end{table}

We perform ablation study experiment on image reconstruction and semantic manipulation tasks on the RAVDESS-12 Dataset. Quantitative comparisons of GAN inversion are shown in Tab.~\ref{table:ablation_inversion}. We can see that every variant outperforms the baseline on all metrics. This indicates both two components contribute to the GAN inversion performance. Meanwhile, variant~($w/o$ MAC) performs better than the variant~($w/o$ ICC), this indicates that the inversion consistency brought by consecutive images contributes more for the GAN inversion task. In Tab.~\ref{table:ablation_semantic} of semantic editing, we observe a different situation. We can see the variant~($w/o$ ICC) performs better than the variant~($w/o$ MAC), this reveals that mutually accessible constraint confines the inverted latent codes to stay in the editable domain. The above two evaluations show that our two constraints work very well following our design principles.

\input{ablation_sem}

We show the results of different variants in Fig.~\ref{fig:abl_semantic} by changing the ``age'' attribute. We can see that the baseline and variant~($w/o$ MAC) entangles with glasses, showing that concentrating only on reconstruction fidelity lacks editability of the inverted codes. In contrast, variant~($w/o$ ICC) and our final result can successfully modify the ``age'' attribute, revealing the strong regularization power of our designed mutually accessible constraint.

% each variant has the noticeable artifacts, and our final results can eliminate them efficiently by cooperating linear transformations with dense correspondence.

%% file: inversion_quality.tex
\begin{figure}[t]
   \centering
    \subfloat[Original]{
    \begin{minipage}{0.14\linewidth}
     \includegraphics[width=\linewidth]{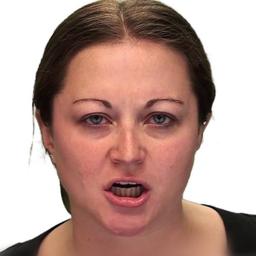}
     \includegraphics[width=\linewidth]{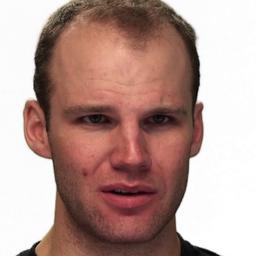}
     \includegraphics[width=\linewidth]{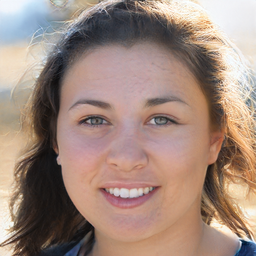}
     \includegraphics[width=\linewidth]{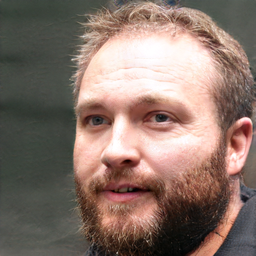}
     \end{minipage}
     }
   \hspace{-2.8mm}
   \subfloat[I2S]{
     \begin{minipage}{0.14\linewidth}
     \includegraphics[width=\linewidth]{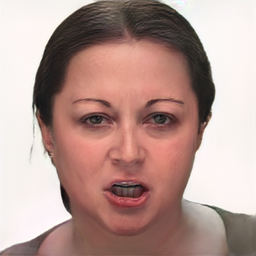}
     \includegraphics[width=\linewidth]{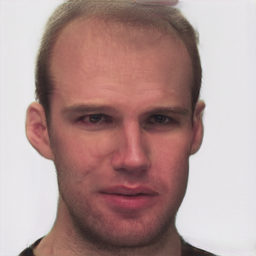}
     \includegraphics[width=\linewidth]{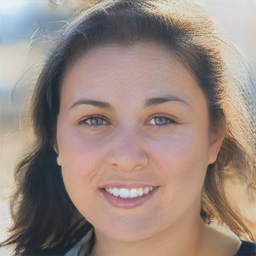}
     \includegraphics[width=\linewidth]{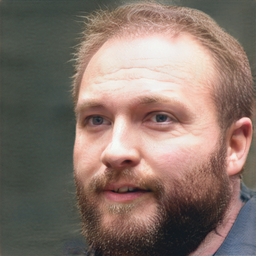}
     \end{minipage}
     }
   \hspace{-2.8mm}
   \subfloat[pSp]{
     \begin{minipage}{0.14\linewidth}
     \includegraphics[width=\linewidth]{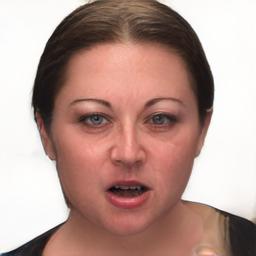}
     \includegraphics[width=\linewidth]{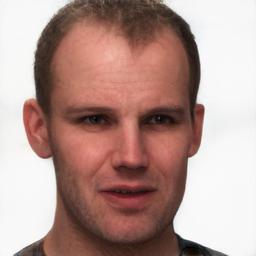}
     \includegraphics[width=\linewidth]{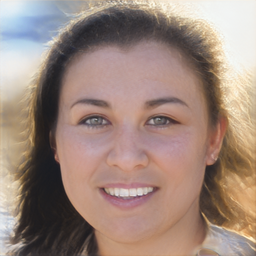}
     \includegraphics[width=\linewidth]{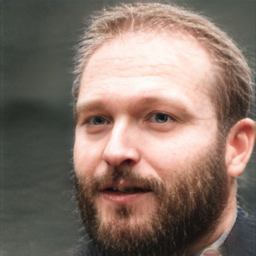}
     \end{minipage}
     }
   \hspace{-2.8mm}
   \subfloat[InD]{
     \begin{minipage}{0.14\linewidth}
     \includegraphics[width=\linewidth]{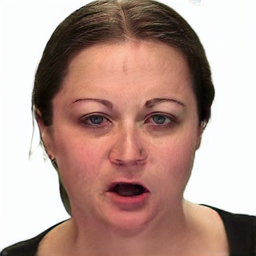}
     \includegraphics[width=\linewidth]{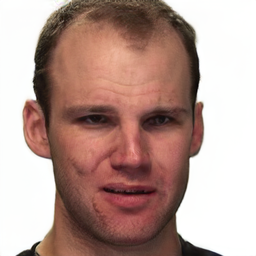}
     \includegraphics[width=\linewidth]{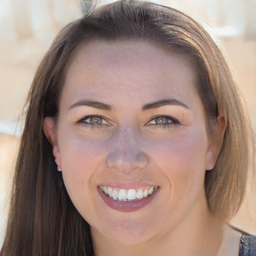}
     \includegraphics[width=\linewidth]{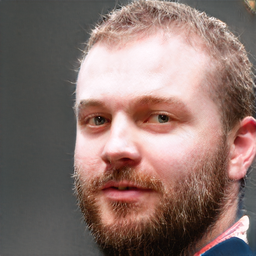}
     \end{minipage}
     }
   \hspace{-2.8mm}
    \subfloat[Ours]{
     \begin{minipage}{0.14\linewidth}
     \includegraphics[width=\linewidth]{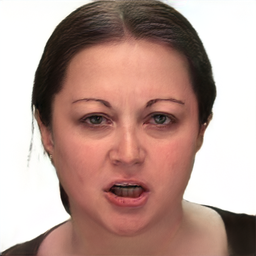}
     \includegraphics[width=\linewidth]{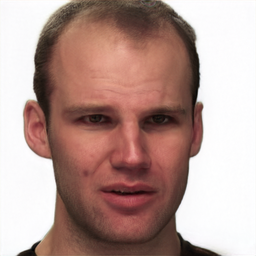}
     \includegraphics[width=\linewidth]{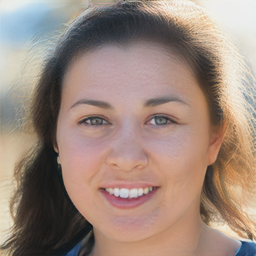}
     \includegraphics[width=\linewidth]{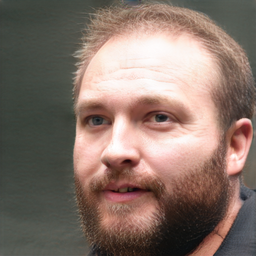}
     \end{minipage}
     }
   % \hspace{-.8mm}
   \hspace{-1.5mm}
   \rotatebox[origin=c]{90}{\rev{- - - - - - - - - - - - - - - - - - - - - - - -}}
   \hspace{-3mm}
   \subfloat[I2S++]{
     \begin{minipage}{0.14\linewidth}
     \includegraphics[width=\linewidth]{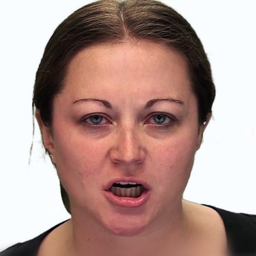}
     \includegraphics[width=\linewidth]{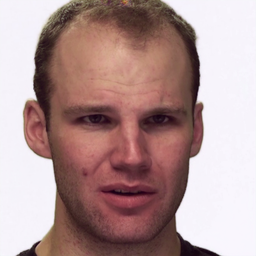}
     \includegraphics[width=\linewidth]{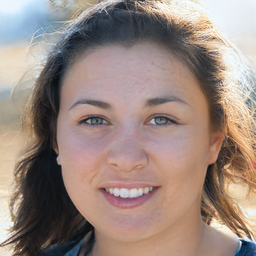}
     \includegraphics[width=\linewidth]{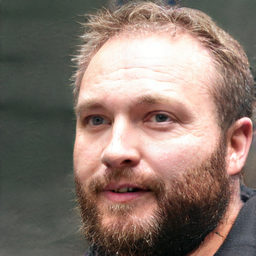}
     \end{minipage}
    }
   \hspace{-2.8mm}
   \subfloat[Ours++]{
     \begin{minipage}{0.14\linewidth}
     \includegraphics[width=\linewidth]{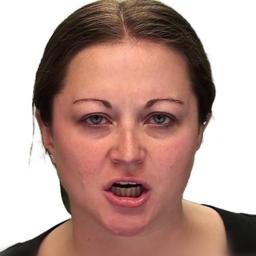}
     \includegraphics[width=\linewidth]{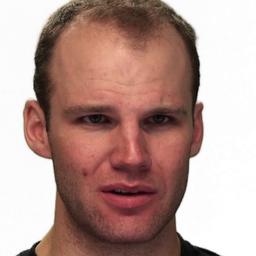}
     \includegraphics[width=\linewidth]{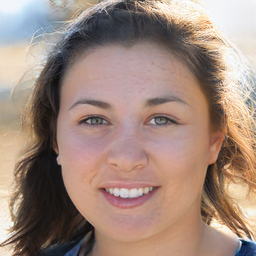}
     \includegraphics[width=\linewidth]{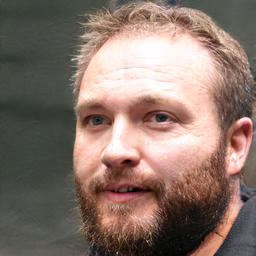}
     \end{minipage}
     }

\vspace{-3mm}\caption{Qualitative comparison on image reconstruction. Compared with the works that optimized in the~$\mathcal{W+}$ space (left part), our method can reconstruct the most faithful appearances. Involving the~$\mathcal{N}$ space largely improves reconstruction (right part), but Ours++ show better color preservation than I2S++ (second row).}
\label{fig:reconstruction_exp}\vspace{-4mm}
\end{figure}

%% file: semantic_editing1.tex
 % %!TeX root = egpaper_for_review.tex

\begin{figure}[t]
   \centering
    \hspace{-3mm}
    % \rotatebox[origin=c]{90}{Ours \hspace{15mm} InD \hspace{15mm} pSp \hspace{15mm} I2S}
    \subfloat[Input]{\begin{minipage}{0.142\linewidth}
     \fcolorbox{white}{red}{\includegraphics[width=\linewidth]{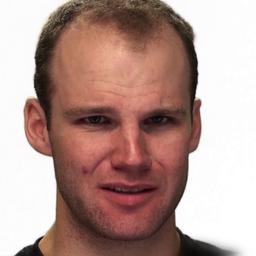}}
     \fcolorbox{white}{white}{\includegraphics[width=\linewidth]{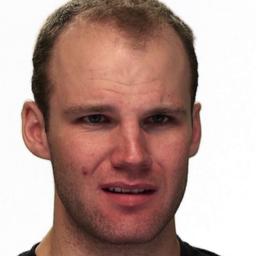}}
     \fcolorbox{white}{white}{\includegraphics[width=\linewidth]{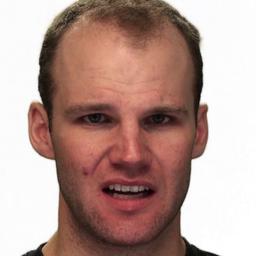}}
     \fcolorbox{white}{red}{\includegraphics[width=\linewidth]{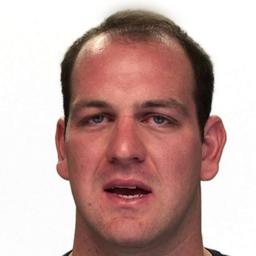}}
     \fcolorbox{white}{white}{\includegraphics[width=\linewidth]{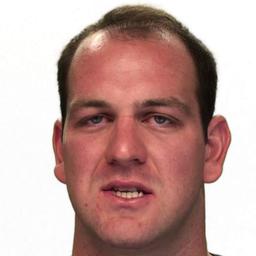}}
     \fcolorbox{white}{white}{\includegraphics[width=\linewidth]{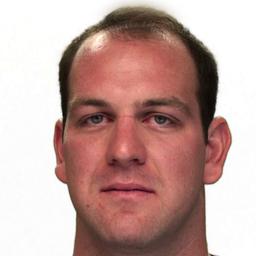}}
     \end{minipage}}
   %\hspace{-2.8mm}
    \subfloat[I2S]{\label{fig:semantic_exp_-p}
     \begin{minipage}{0.142\linewidth}
     \fcolorbox{white}{white}{\includegraphics[width=\linewidth]{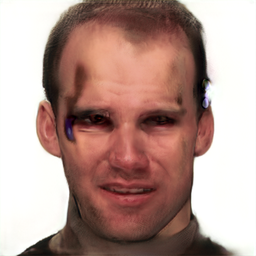}}
     \fcolorbox{white}{white}{\includegraphics[width=\linewidth]{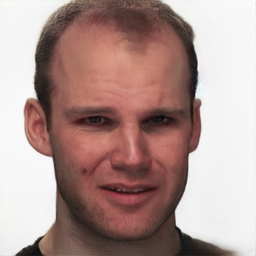}}
     \fcolorbox{white}{white}{\includegraphics[width=\linewidth]{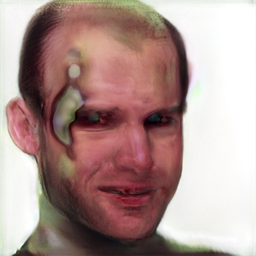}}
     \fcolorbox{white}{white}{\includegraphics[width=\linewidth]{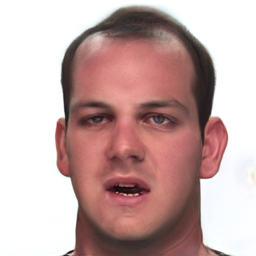}}
     \fcolorbox{white}{white}{\includegraphics[width=\linewidth]{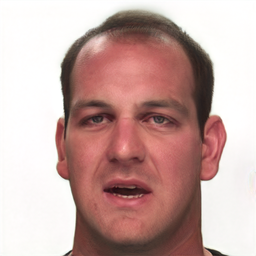}}
     \fcolorbox{white}{white}{\includegraphics[width=\linewidth]{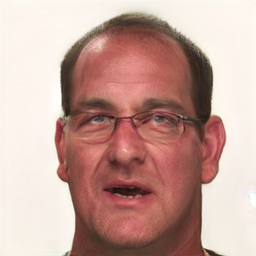}}
     \end{minipage}
     }
   \hspace{-2.8mm}
    \subfloat[Ind]{\label{fig:semantic_exp_+p}
     \begin{minipage}{0.142\linewidth}
     \fcolorbox{white}{white}{\includegraphics[width=\linewidth]{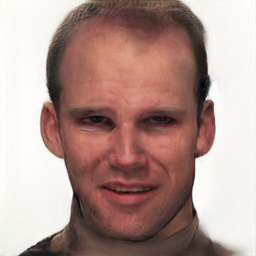}}
     \fcolorbox{white}{white}{\includegraphics[width=\linewidth]{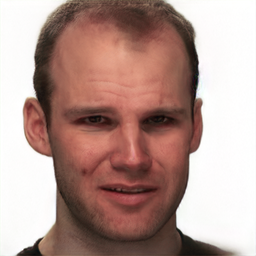}}
     \fcolorbox{white}{white}{\includegraphics[width=\linewidth]{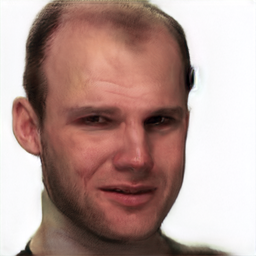}}
     \fcolorbox{white}{white}{\includegraphics[width=\linewidth]{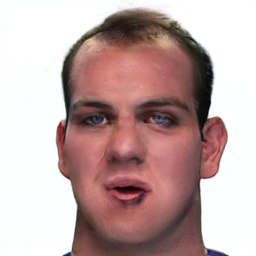}}
     \fcolorbox{white}{white}{\includegraphics[width=\linewidth]{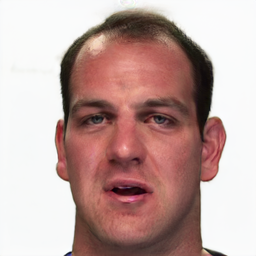}}
     \fcolorbox{white}{white}{\includegraphics[width=\linewidth]{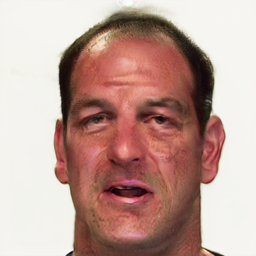}}
     \end{minipage}
     }
    \hspace{-2.8mm}
    \subfloat[pSp]{
     \begin{minipage}{0.142\linewidth}
     \fcolorbox{white}{white}{\includegraphics[width=\linewidth]{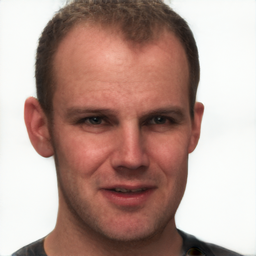}}
     \fcolorbox{white}{white}{\includegraphics[width=\linewidth]{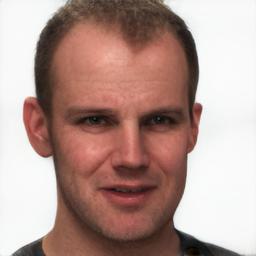}}
     \fcolorbox{white}{white}{\includegraphics[width=\linewidth]{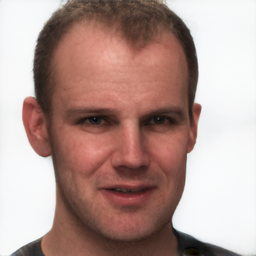}}
     \fcolorbox{white}{white}{\includegraphics[width=\linewidth]{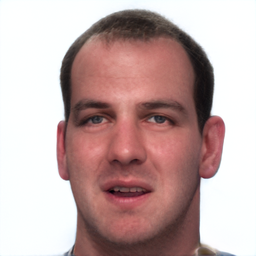}}
     \fcolorbox{white}{white}{\includegraphics[width=\linewidth]{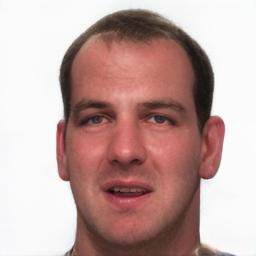}}
     \fcolorbox{white}{white}{\includegraphics[width=\linewidth]{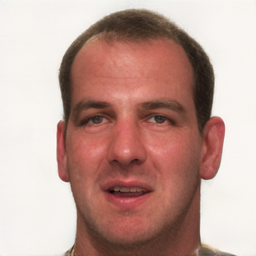}}
     \end{minipage}
     }
     \hspace{-2.8mm}
     \subfloat[Ours]{\label{fig:semantic_exp_g}
     \begin{minipage}{0.142\linewidth}
     \fcolorbox{white}{white}{\includegraphics[width=\linewidth]{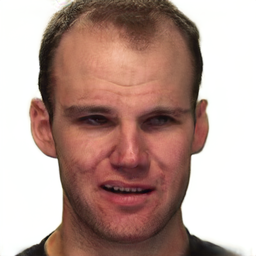}}
     \fcolorbox{white}{white}{\includegraphics[width=\linewidth]{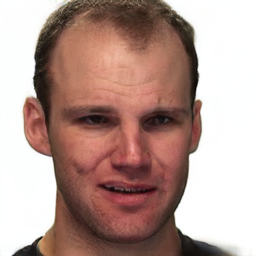}}
     \fcolorbox{white}{white}{\includegraphics[width=\linewidth]{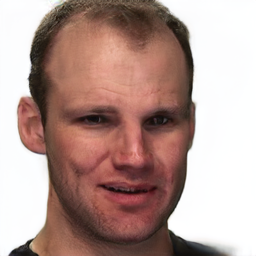}}
     \fcolorbox{white}{white}{\includegraphics[width=\linewidth]{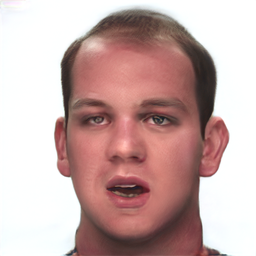}}
     \fcolorbox{white}{white}{\includegraphics[width=\linewidth]{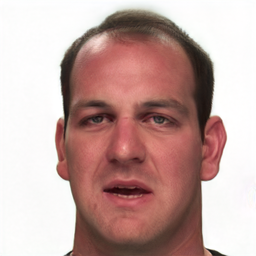}}
     \fcolorbox{white}{white}{\includegraphics[width=\linewidth]{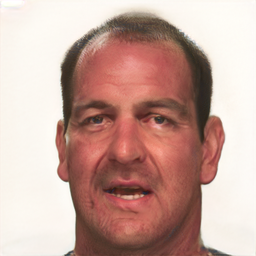}}
    \end{minipage}
     }
   \hspace{-1.5mm}
   \rotatebox[origin=c]{90}{\rev{- - - - - - - - - - - - - - - - - - - - - - - - - - - - - - - - - - - -}}
   \hspace{-3mm}
    \subfloat[I2S++]{\label{fig:semantic_exp_-p++}
     \begin{minipage}{0.142\linewidth}
     \fcolorbox{white}{white}{\includegraphics[width=\linewidth]{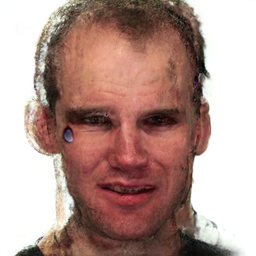}}
     \fcolorbox{white}{white}{\includegraphics[width=\linewidth]{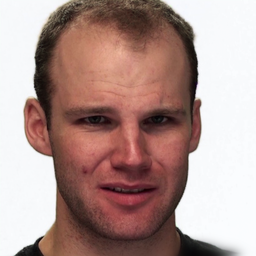}}
     \fcolorbox{white}{white}{\includegraphics[width=\linewidth]{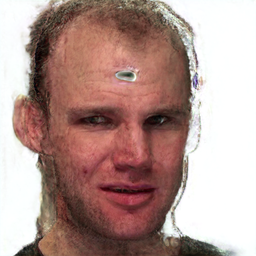}}
     \fcolorbox{white}{white}{\includegraphics[width=\linewidth]{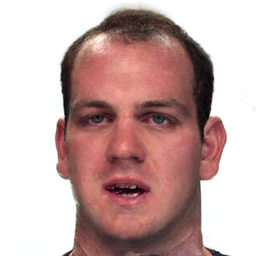}}
     \fcolorbox{white}{white}{\includegraphics[width=\linewidth]{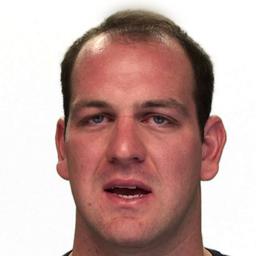}}
     \fcolorbox{white}{white}{\includegraphics[width=\linewidth]{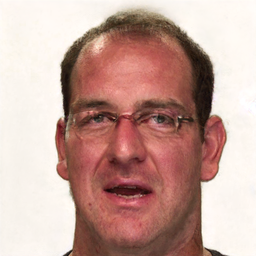}}
     \end{minipage}
     }
   \hspace{-2.8mm}
   \subfloat[Ours++]{\label{fig:semantic_exp_-p++2}
     \begin{minipage}{0.142\linewidth}
     \fcolorbox{white}{white}{\includegraphics[width=\linewidth]{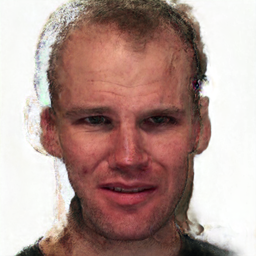}}
     \fcolorbox{white}{white}{\includegraphics[width=\linewidth]{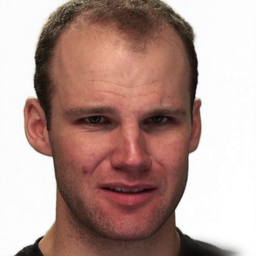}}
     \fcolorbox{white}{white}{\includegraphics[width=\linewidth]{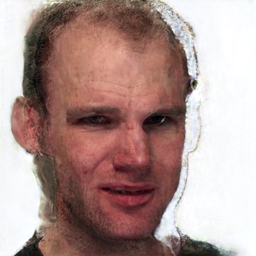}}
     \fcolorbox{white}{white}{\includegraphics[width=\linewidth]{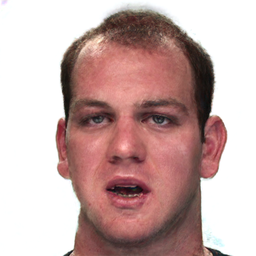}}
     \fcolorbox{white}{white}{\includegraphics[width=\linewidth]{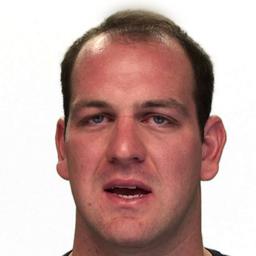}}
     \fcolorbox{white}{white}{\includegraphics[width=\linewidth]{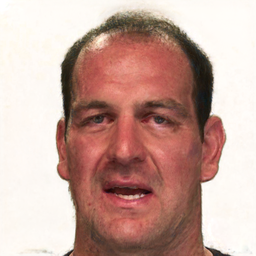}}
     \end{minipage}
     }
   \hspace{-3mm}
    \rotatebox[origin=c]{90}{  \textbf{+}  \hspace{5mm} \textbf{Age}   \hspace{5mm} \textbf{--} \hspace{10mm} \textbf{+}  \hspace{5mm} \textbf{Pose} \hspace{5mm}  \textbf{--}}
\vspace{-3mm}\caption{Qualitative comparison on semantic editing with Pose and Age attributes on the real RAVDESS-12 Dataset. Images marked by red boxes are the reconstructed targets, and images in the middle row of each example are the inversion results. We can tell that our method can support more favorable semantic editing.}
\label{fig:semantic_exp}\vspace{-2mm}
\end{figure}

%% file: semantic_editing2.tex
 \begin{figure}[t]
   \centering
    \hspace{-5mm}
    \rotatebox[origin=c]{90}{  \textbf{+}  \hspace{5mm} \textbf{Age}   \hspace{5mm} \textbf{--} \hspace{10mm} \textbf{+}  \hspace{5mm} \textbf{Smile} \hspace{5mm}  \textbf{--}}
    \subfloat[GT]{
    \begin{minipage}{0.195\linewidth}
     \includegraphics[width=\linewidth]{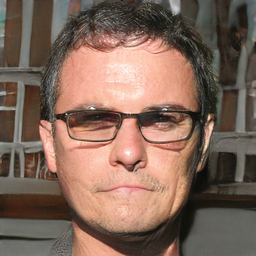}
     \includegraphics[width=\linewidth]{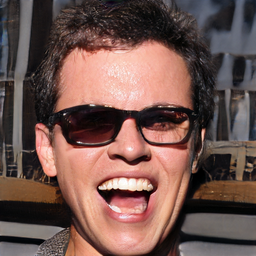}
     \includegraphics[width=\linewidth]{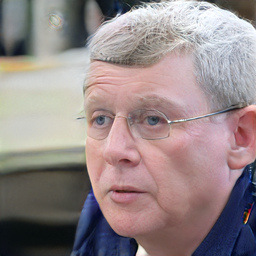}
     \includegraphics[width=\linewidth]{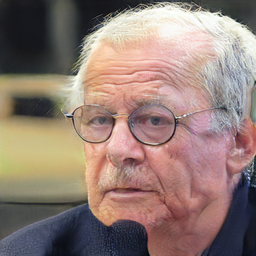}
     \end{minipage}
     }
   \hspace{-2.8mm}
   \subfloat[I2S]{
     \begin{minipage}{0.195\linewidth}
     \includegraphics[width=\linewidth]{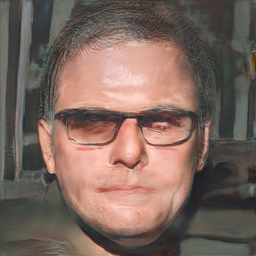}
     \includegraphics[width=\linewidth]{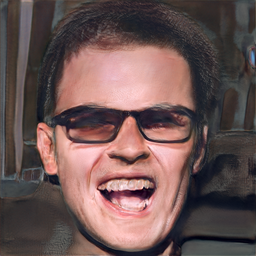}
     \includegraphics[width=\linewidth]{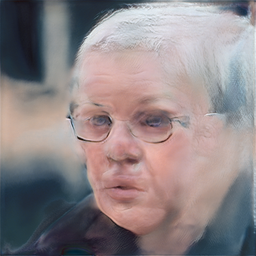}
     \includegraphics[width=\linewidth]{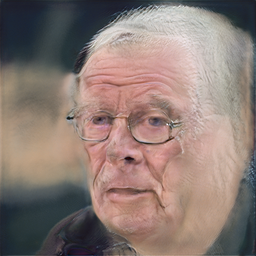}
     \end{minipage}
     }
   \hspace{-2.8mm}
    \subfloat[Ind]{
     \begin{minipage}{0.195\linewidth}
     \includegraphics[width=\linewidth]{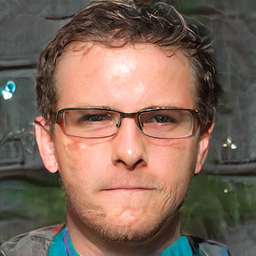}
     \includegraphics[width=\linewidth]{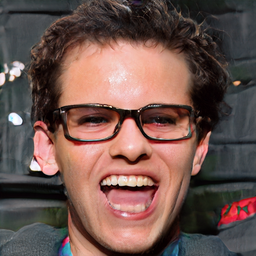}
     \includegraphics[width=\linewidth]{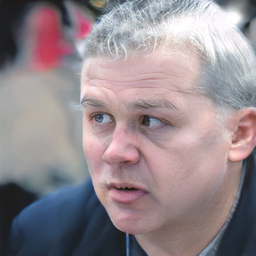}
     \includegraphics[width=\linewidth]{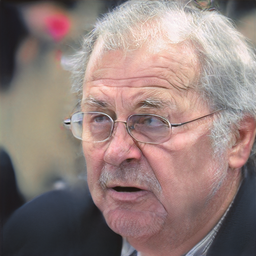}

     \end{minipage}
     }
   \hspace{-2.8mm}
    \subfloat[pSp]{
     \begin{minipage}{0.195\linewidth}
     \includegraphics[width=\linewidth]{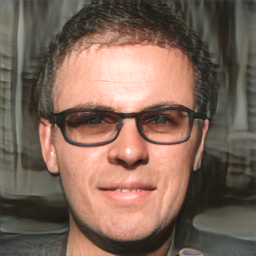}
     \includegraphics[width=\linewidth]{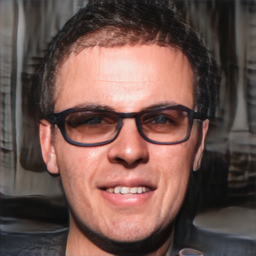}
     \includegraphics[width=\linewidth]{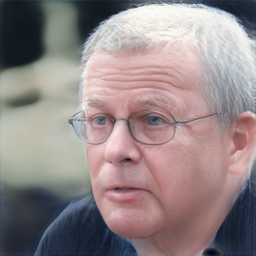}
     \includegraphics[width=\linewidth]{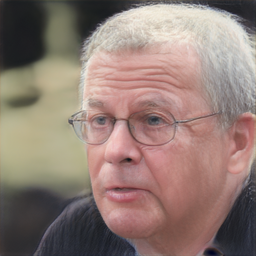}
     \end{minipage}
     }
   \hspace{-2.8mm}
   \subfloat[Ours]{
     \begin{minipage}{0.195\linewidth}
     \includegraphics[width=\linewidth]{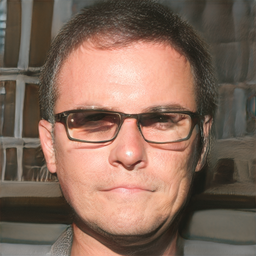}
     \includegraphics[width=\linewidth]{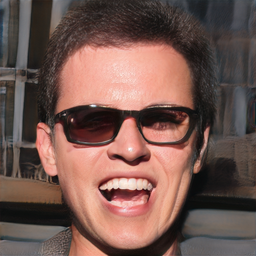}
     \includegraphics[width=\linewidth]{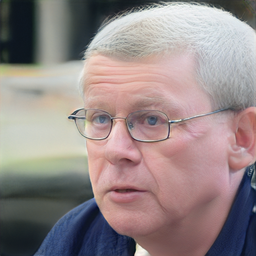}
     \includegraphics[width=\linewidth]{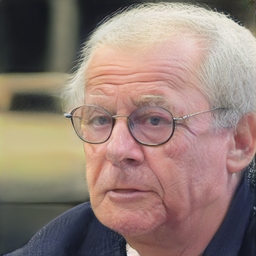}
     \end{minipage}
     }
   \hspace{-2.8mm}
\vspace{-3mm}\caption{Qualitative comparison on semantic editing with Smile and Age attributes on the Synthesized Dataset. It's noticed that we force the input sequence to contain different semantic changes from its corresponding editing direction. the first sequence contains the semantic changes with ``\emph{gender}'' for optimization and ``\emph{smile}'' for semantic edit, and the second is ``\emph{pose}'' for optimization while ``\emph{age}'' for the edit testing. And our edited results are more similar with the ground truths.}
\label{fig:semantic_exp_g}\vspace{-3mm}
\end{figure}

%% file: image_morphing.tex
\begin{figure}[t]
   \centering
    \hspace{-5mm}
    \rotatebox[origin=c]{90}{ Ours \hspace{10mm} InD \hspace{10mm} pSp \hspace{10mm} I2S}
    % \hspace{-2mm}
    %\hspace{-2.8mm}
    {%\label{fig:semantic_exp_-p}
     \begin{minipage}{0.195\linewidth}
     \includegraphics[width=\linewidth]{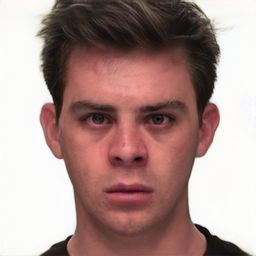}
     \includegraphics[width=\linewidth]{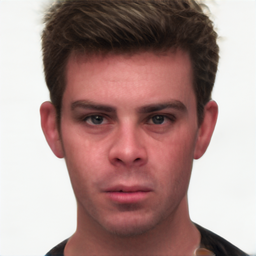}
     \includegraphics[width=\linewidth]{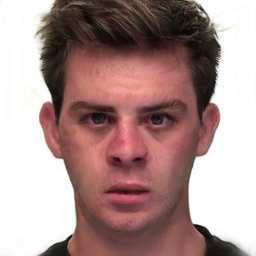}
     \includegraphics[width=\linewidth]{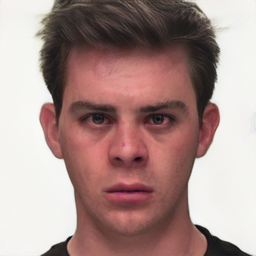}
     \end{minipage}
     }
    \hspace{-2.8mm}
    {%\label{fig:semantic_exp_-p}
     \begin{minipage}{0.195\linewidth}
     \includegraphics[width=\linewidth]{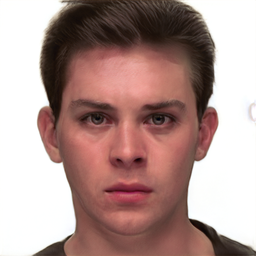}
     \includegraphics[width=\linewidth]{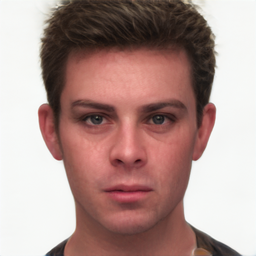}
     \includegraphics[width=\linewidth]{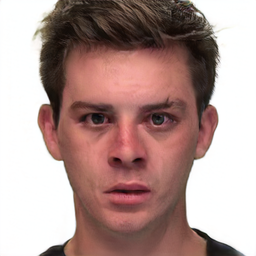}
     \includegraphics[width=\linewidth]{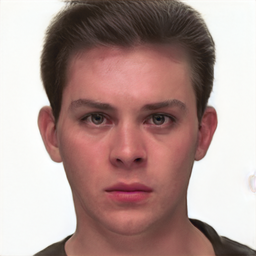}
     \end{minipage}
     }
    \hspace{-2.8mm}
    {%\label{fig:semantic_exp_-p}
     \begin{minipage}{0.195\linewidth}
     \includegraphics[width=\linewidth]{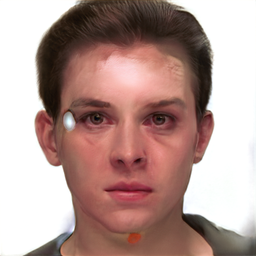}
     \includegraphics[width=\linewidth]{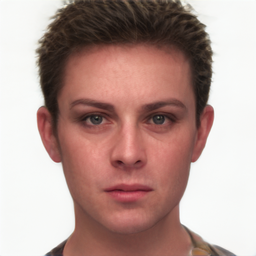}
     \includegraphics[width=\linewidth]{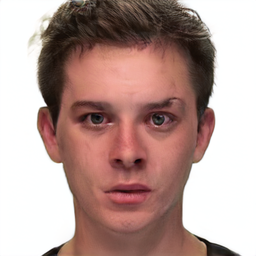}
     \includegraphics[width=\linewidth]{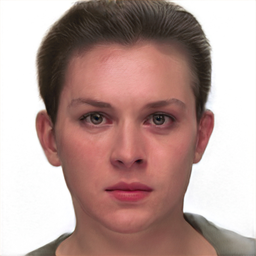}
     \end{minipage}
     }
    \hspace{-2.8mm}
    {%\label{fig:semantic_exp_-p}
     \begin{minipage}{0.195\linewidth}
     \includegraphics[width=\linewidth]{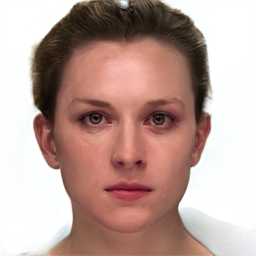}
     \includegraphics[width=\linewidth]{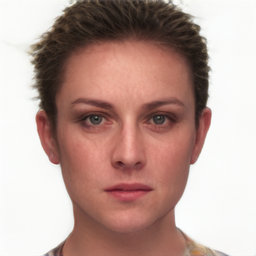}
     \includegraphics[width=\linewidth]{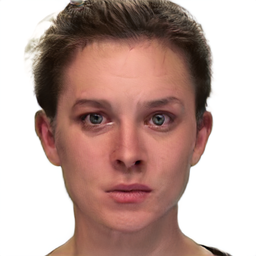}
     \includegraphics[width=\linewidth]{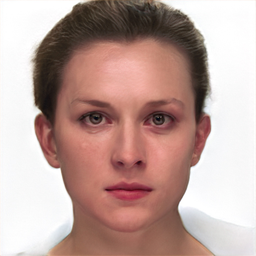}
     \end{minipage}
     }
    \hspace{-2.8mm}
    {%\label{fig:semantic_exp_-p}
     \begin{minipage}{0.195\linewidth}
     \includegraphics[width=\linewidth]{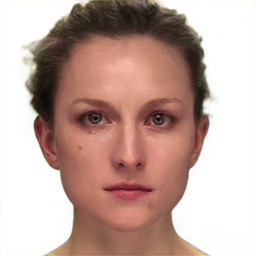}
     \includegraphics[width=\linewidth]{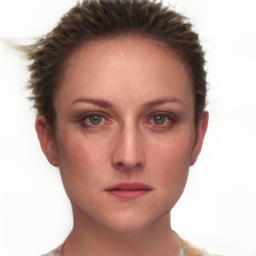}
     \includegraphics[width=\linewidth]{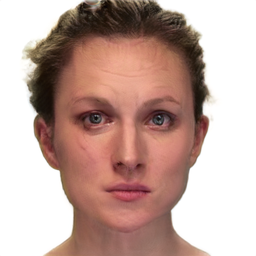}
     \includegraphics[width=\linewidth]{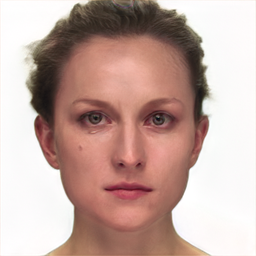}
     \end{minipage}
     }
    \rotatebox[origin=c]{0}{Inverted A$ \hspace{3mm} \longleftarrow$  \hspace{8mm} Morphing  \hspace{8mm} $\longrightarrow \hspace{3mm}$Inverted B}

\vspace{-3mm}\caption{Qualitative comparison on image morphing task. We can see that our result present a continuous process and the morphing faces are realistic.}
\label{fig:image_morphing}\vspace{-5mm}
\end{figure}

%% file: semantic_transfer.tex
\begin{figure*}[t]
   \centering
    % \setlength{\fboxrule}{0.5pt}
    % \setlength{\fboxsep}{0cm}
    %\rotatebox[origin=c]{90}{  \textbf{+}  \hspace{5mm} \textbf{Age}   \hspace{5mm} \textbf{--} \hspace{10mm} \textbf{+}  \hspace{5mm} \textbf{Smile} \hspace{5mm}  \textbf{--}}
    {
    \begin{minipage}{0.09\linewidth}
     \fcolorbox{white}{white}{\includegraphics[width=\linewidth]{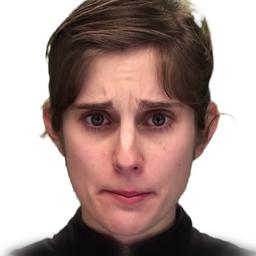}}
     \fcolorbox{white}{white}{\includegraphics[width=\linewidth]{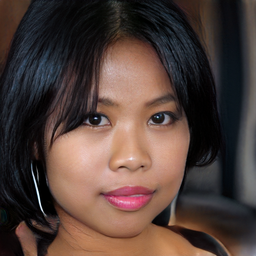}}
     \fcolorbox{white}{white}{\includegraphics[width=\linewidth]{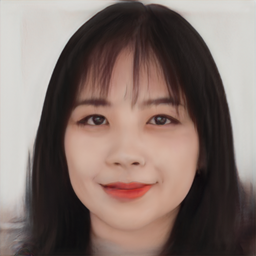}}
     \fcolorbox{white}{white}{\includegraphics[width=\linewidth]{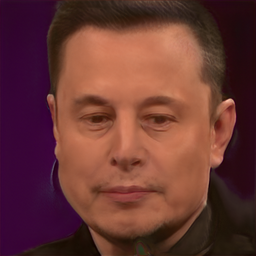}}
     \end{minipage}
     }
   \hspace{-2.8mm}
   {
     \begin{minipage}{0.09\linewidth}
     \fcolorbox{white}{white}{\includegraphics[width=\linewidth]{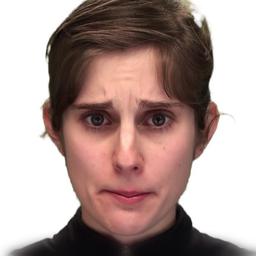}}
     \fcolorbox{white}{white}{\includegraphics[width=\linewidth]{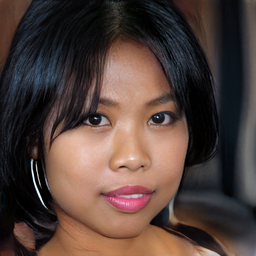}}
     \fcolorbox{white}{white}{\includegraphics[width=\linewidth]{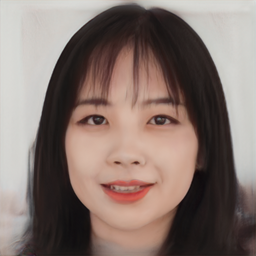}}
     \fcolorbox{white}{white}{\includegraphics[width=\linewidth]{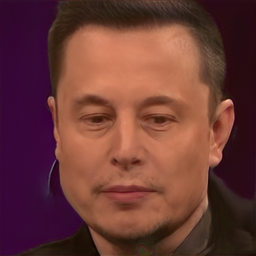}}
     \end{minipage}
     }
   \hspace{-2.8mm}
   {
     \begin{minipage}{0.09\linewidth}
     \fcolorbox{white}{white}{\includegraphics[width=\linewidth]{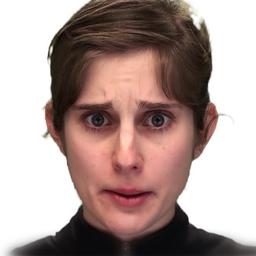}}
     \fcolorbox{white}{red}{\includegraphics[width=\linewidth]{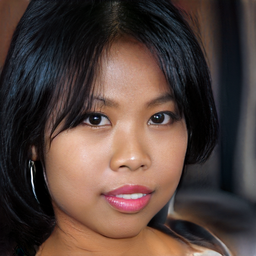}}
     \fcolorbox{white}{red}{\includegraphics[width=\linewidth]{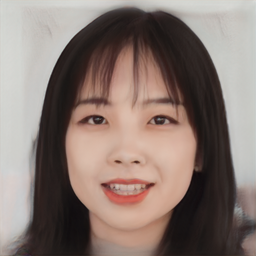}}
     \fcolorbox{white}{red}{\includegraphics[width=\linewidth]{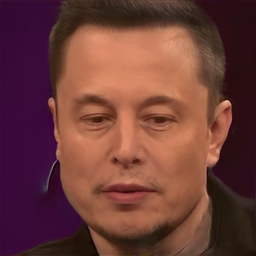}}
     \end{minipage}
     }
   \hspace{-2.8mm}
    {
     \begin{minipage}{0.09\linewidth}
     \fcolorbox{white}{white}{\includegraphics[width=\linewidth]{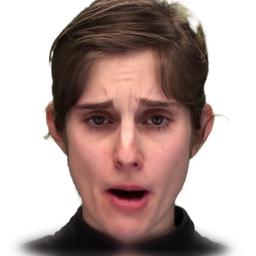}}
     \fcolorbox{white}{white}{\includegraphics[width=\linewidth]{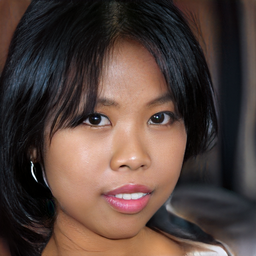}}
     \fcolorbox{white}{white}{\includegraphics[width=\linewidth]{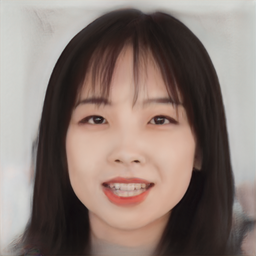}}
     \fcolorbox{white}{white}{\includegraphics[width=\linewidth]{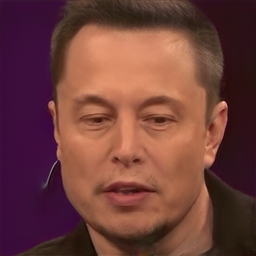}}
     \end{minipage}
     }
   \hspace{-2.8mm}
    {
     \begin{minipage}{0.09\linewidth}
     \fcolorbox{white}{white}{\includegraphics[width=\linewidth]{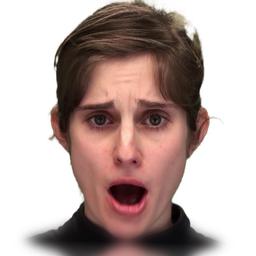}}
     \fcolorbox{white}{white}{\includegraphics[width=\linewidth]{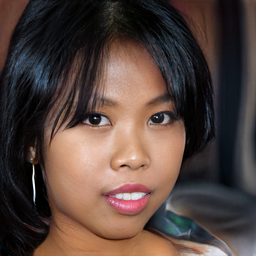}}
     \fcolorbox{white}{white}{\includegraphics[width=\linewidth]{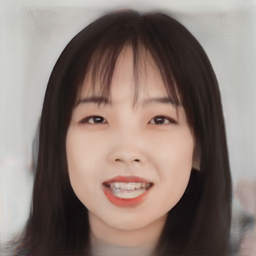}}
     \fcolorbox{white}{white}{\includegraphics[width=\linewidth]{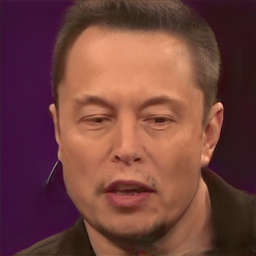}}
     \end{minipage}
     }
    {
     \begin{minipage}{0.09\linewidth}
     \fcolorbox{white}{white}{\includegraphics[width=\linewidth]{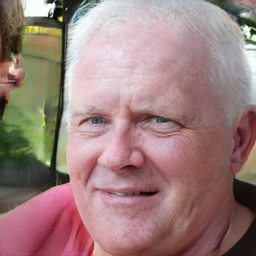}}
     \fcolorbox{white}{red}{\includegraphics[width=\linewidth]{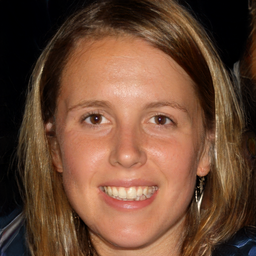}}
     \fcolorbox{white}{red}{\includegraphics[width=\linewidth]{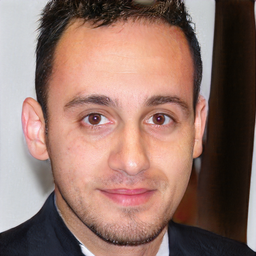}}
     \fcolorbox{white}{red}{\includegraphics[width=\linewidth]{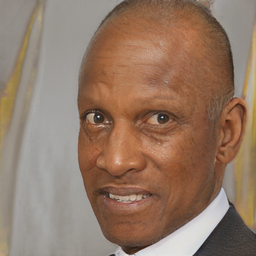}}
     \end{minipage}
     }
   \hspace{-2.8mm}
   {
     \begin{minipage}{0.09\linewidth}
     \fcolorbox{white}{white}{\includegraphics[width=\linewidth]{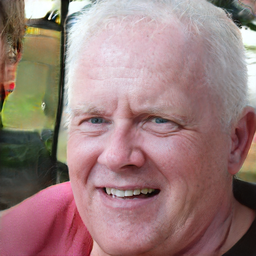}}
     \fcolorbox{white}{white}{\includegraphics[width=\linewidth]{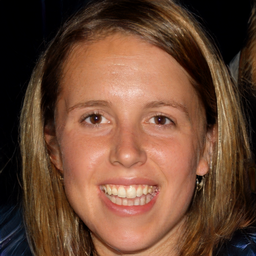}}
     \fcolorbox{white}{white}{\includegraphics[width=\linewidth]{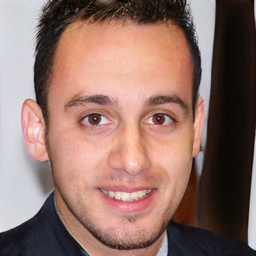}}
     \fcolorbox{white}{white}{\includegraphics[width=\linewidth]{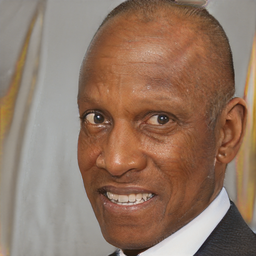}}
     \end{minipage}
     }
   \hspace{-2.8mm}
    {
    \begin{minipage}{0.09\linewidth}
     \fcolorbox{white}{white}{\includegraphics[width=\linewidth]{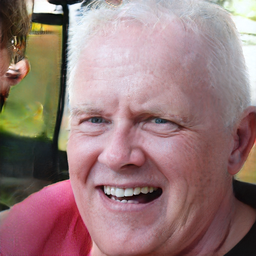}}
     \fcolorbox{white}{white}{\includegraphics[width=\linewidth]{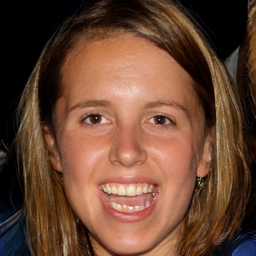}}
     \fcolorbox{white}{white}{\includegraphics[width=\linewidth]{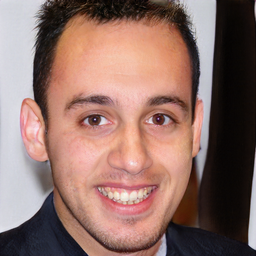}}
     \fcolorbox{white}{white}{\includegraphics[width=\linewidth]{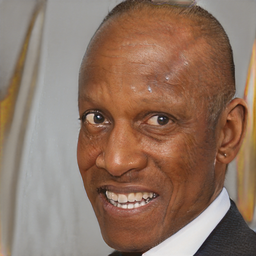}}
     \end{minipage}
     }
   \hspace{-2.8mm}
    {
     \begin{minipage}{0.09\linewidth}
     \fcolorbox{white}{white}{\includegraphics[width=\linewidth]{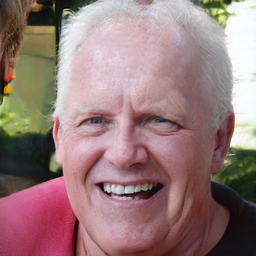}}
     \fcolorbox{white}{white}{\includegraphics[width=\linewidth]{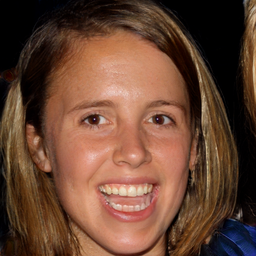}}
     \fcolorbox{white}{white}{\includegraphics[width=\linewidth]{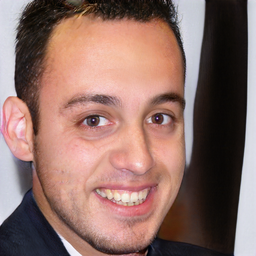}}
     \fcolorbox{white}{white}{\includegraphics[width=\linewidth]{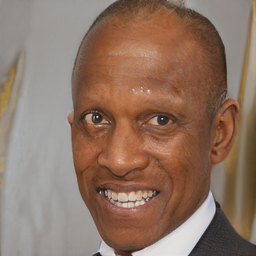}}
     \end{minipage}
     }
   \hspace{-2.8mm}
    {
     \begin{minipage}{0.09\linewidth}
     \fcolorbox{white}{white}{\includegraphics[width=\linewidth]{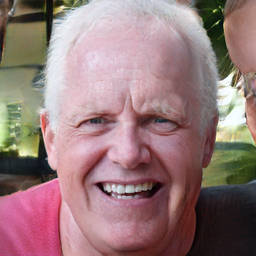}}
     \fcolorbox{white}{white}{\includegraphics[width=\linewidth]{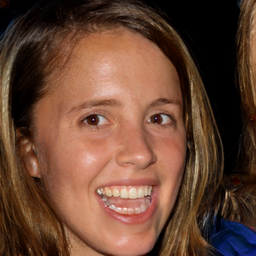}}
     \fcolorbox{white}{white}{\includegraphics[width=\linewidth]{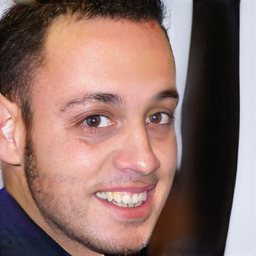}}
     \fcolorbox{white}{white}{\includegraphics[width=\linewidth]{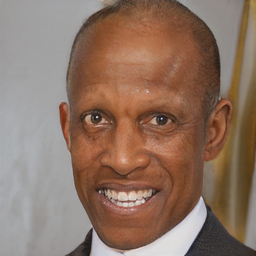}}
     \end{minipage}
     }

\vspace{-3mm}\caption{Our unsupervisedly acquired direction~$\bold{\mathop{n}\limits ^{\rightarrow}}$~from consecutive images can be used for transferring the semantics. The first row is the input set that is regarded as reference, and the images in red boxes are the target faces. We can transfer the semantic changes of the reference to the target faces, even with more than one attribute changed.}
\label{fig:semantic_trans}\vspace{-6mm}
\end{figure*} 

%% file: ablation_sem.tex
%!TeX root = egpaper_for_review.tex

\begin{figure}[t]
    \centering
    \vspace{-3mm}
    \hspace{-5mm}
    % \rotatebox[origin=c]{90}{\hspace{-3mm} $\alpha>0$ \hspace{10mm} $\alpha=0$ \hspace{10mm} $\alpha<0$}
    \rotatebox[origin=c]{90}{  \textbf{+}  \hspace{5mm} \textbf{Age}   \hspace{5mm} \textbf{--}}
    \subfloat[Baseline]{\label{fig:abl_exp_-a}
     \begin{minipage}{0.24\linewidth}
     \includegraphics[width=\linewidth]{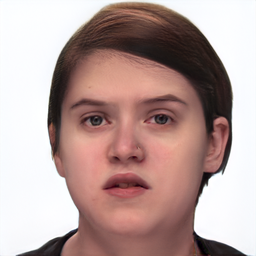}
     \includegraphics[width=\linewidth]{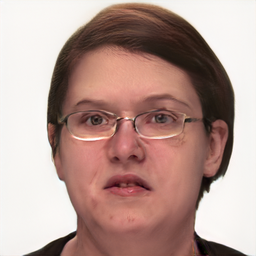}
     \end{minipage}
     }
   \hspace{-2.8mm}
   \subfloat[$w/o$ MAS]{\label{fig:abl_exp_-b}
     \begin{minipage}{0.24\linewidth}
     \includegraphics[width=\linewidth]{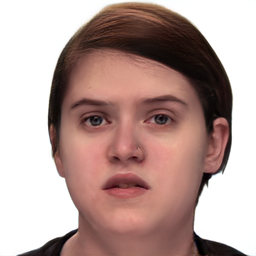}
     \includegraphics[width=\linewidth]{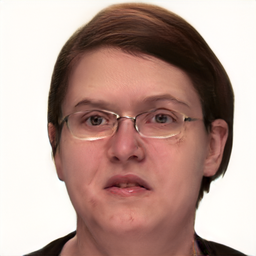}
     \end{minipage}
     }
   \hspace{-2.8mm}
   \subfloat[$w/o$ ICS]{
     \begin{minipage}{0.24\linewidth}
     \includegraphics[width=\linewidth]{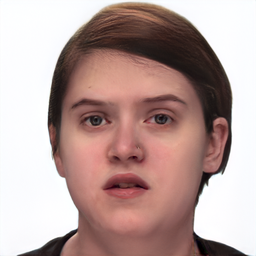}
     \includegraphics[width=\linewidth]{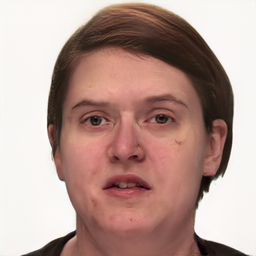}
     \end{minipage}
     }
   \hspace{-2.8mm}
    \subfloat[Ours]{
     \begin{minipage}{0.24\linewidth}
     \includegraphics[width=\linewidth]{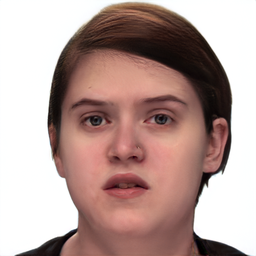}
     \includegraphics[width=\linewidth]{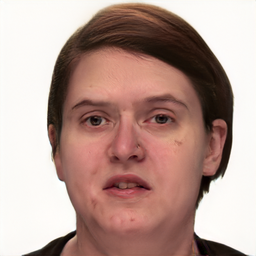}
     \end{minipage}
     }
   \hspace{-2.8mm}
\vspace{-3mm}\caption{Ablation study on semantic editing with two variants and baseline by editing the ``age'' attribute.}
\label{fig:abl_semantic}\vspace{-6mm}
\end{figure}

%% file: 5.conclusion.tex
% !TeX root = egpaper_for_review.tex
\vspace{-2mm}\section{Conclusion}\vspace{-2mm}
In this paper, we propose an alternative GAN inversion method for consecutive images, we formulate consecutive images inversion as a linear combination process in the latent space that ensures editability, and transfer the reconstruction consistency across inputs in the RGB space to guarantee reconstruction fidelity. The experiment results demonstrate the effectiveness in terms of editability and reconstruction fidelity. Besides, we also support various new applications like video-based GAN inversion and unsupervised semantic transfer.

%% file: supp3.tex
\section{Supplemental Results}

\begin{table*}[t]
   \caption{Quantitative evaluations on image reconstruction and semantic editing tasks on the nonlinear dataset.}
       \vspace{-0.6cm}
       \begin{center}
       \setlength{\tabcolsep}{0.2cm}{
       % \resizebox{0.475\textwidth}{!}{
       \begin{tabular}{c|c|c|c|c|c|c|c|c}
           \hline
           \multirow{1}{*}{\diagbox{Metrics}{Methods}}        & \multicolumn{4}{c|}{Image Reconstruction} & \multicolumn{4}{c}{Nonlinear Semantic Edit} \\
           \cline{2-9}
                             &I2S   &pSp   &InD  &Ours  &I2S  &pSp  &InD  &Ours\\
           \hline
           NIQE  $\downarrow$                      &   3.632         &3.439         &3.254      &\textbf{2.997}                                        &   3.940         &3.974          &3.703      &\textbf{3.476}          \\

           FID  $\downarrow$                       &   40.098        &62.932        &77.692      &\textbf{33.692}
                                                   &   48.032        &64.224         &51.607     &\textbf{37.039}         \\

           LPIPS  $\downarrow$                     &   0.252         &0.323          &0.414      &\textbf{0.238}
                                                   &   0.489         &0.522          &0.471      &\textbf{0.403}         \\
           MSE$\downarrow$($\times$e-3)            &   30.767        &79.878        &83.294     &\textbf{25.192}
                                                   &   87.361        &118.285        &99.623     &\textbf{69.449}         \\
           \hline
       \end{tabular}
       }
       % }
      \end{center}
      \label{table}
\end{table*}

\begin{figure*}[t]
    \centering
    \subfloat[Original]{
    \begin{minipage}{0.18\linewidth}
     \includegraphics[width=\linewidth]{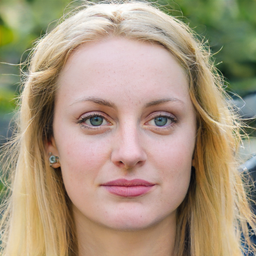}
     \includegraphics[width=\linewidth]{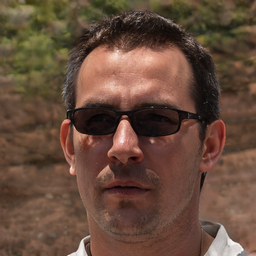}
     \includegraphics[width=\linewidth]{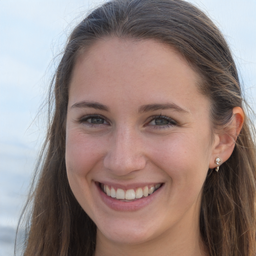}
     \includegraphics[width=\linewidth]{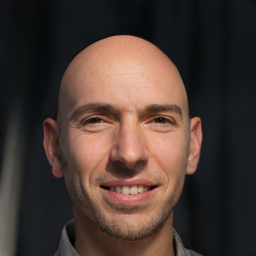}
     \end{minipage}
     }
   \hspace{-2.8mm}
   \subfloat[I2S]{
     \begin{minipage}{0.18\linewidth}
     \includegraphics[width=\linewidth]{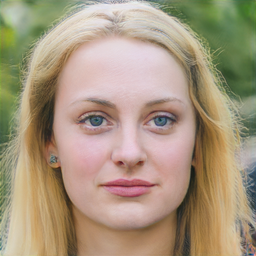}
     \includegraphics[width=\linewidth]{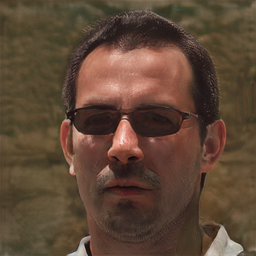}
     \includegraphics[width=\linewidth]{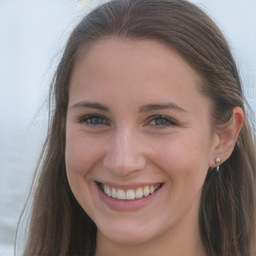}
     \includegraphics[width=\linewidth]{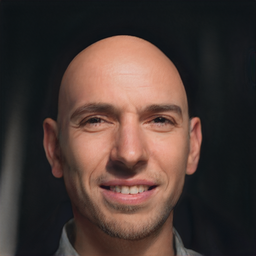}
     \end{minipage}
     }
   \hspace{-2.8mm}
   \subfloat[pSp]{
     \begin{minipage}{0.18\linewidth}
     \includegraphics[width=\linewidth]{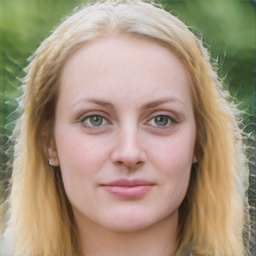}
     \includegraphics[width=\linewidth]{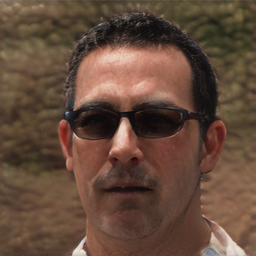}
     \includegraphics[width=\linewidth]{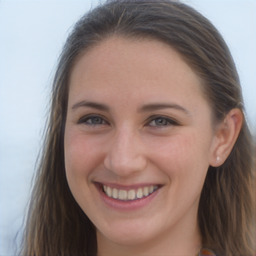}
     \includegraphics[width=\linewidth]{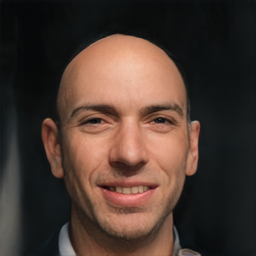}
     \end{minipage}
     }
   \hspace{-2.8mm}
   \subfloat[InD]{
     \begin{minipage}{0.18\linewidth}
     \includegraphics[width=\linewidth]{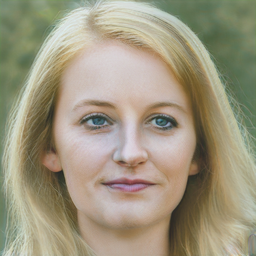}
     \includegraphics[width=\linewidth]{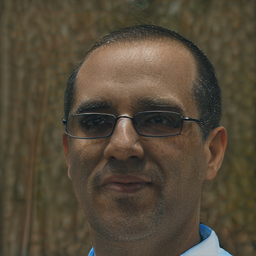}
     \includegraphics[width=\linewidth]{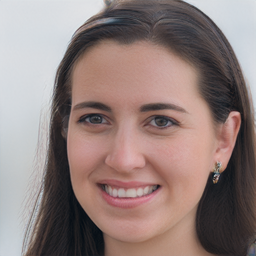}
     \includegraphics[width=\linewidth]{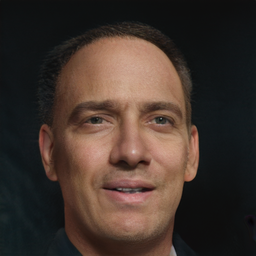}
     \end{minipage}
     }
   \hspace{-2.8mm}
    \subfloat[Ours]{
     \begin{minipage}{0.18\linewidth}
     \includegraphics[width=\linewidth]{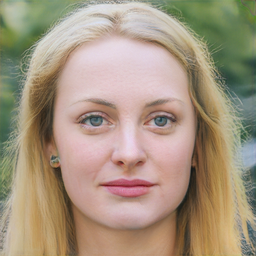}
     \includegraphics[width=\linewidth]{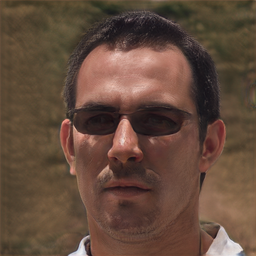}
     \includegraphics[width=\linewidth]{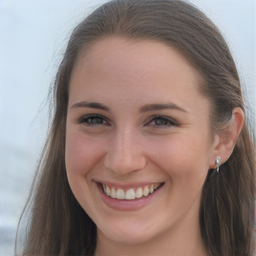}
     \includegraphics[width=\linewidth]{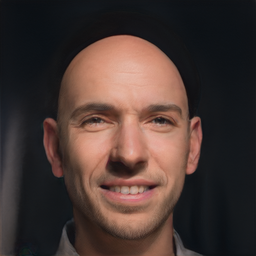}
     \end{minipage}
     }
   
\vspace{-3mm}\caption{Qualitative comparison on image reconstruction on the nonlinear dataset.}
\label{fig_rec}\vspace{-6mm}
\end{figure*}

\begin{figure*}[t]
    \centering
    \subfloat[GT]{
    \begin{minipage}{0.18\linewidth}
     \includegraphics[width=\linewidth]{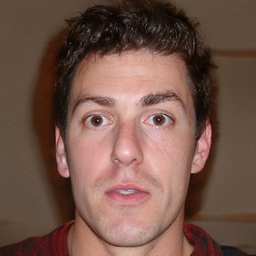}
     \includegraphics[width=\linewidth]{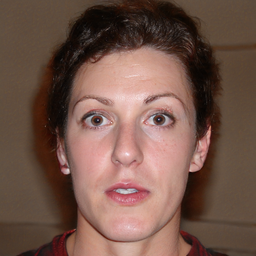}
     \includegraphics[width=\linewidth]{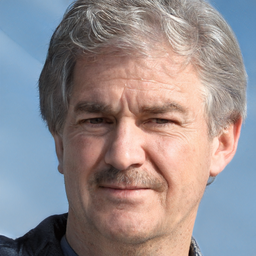}
     \includegraphics[width=\linewidth]{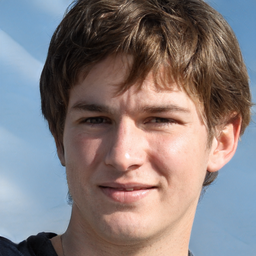}
     \end{minipage}
     }
   \hspace{-2.8mm}
   \subfloat[I2S]{
     \begin{minipage}{0.18\linewidth}
     \includegraphics[width=\linewidth]{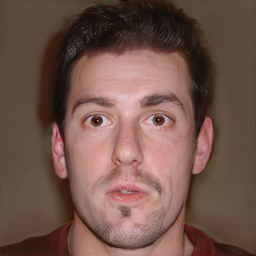}
     \includegraphics[width=\linewidth]{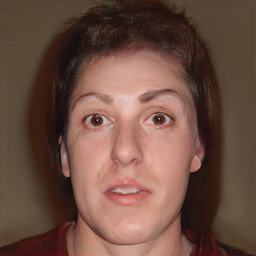}
     \includegraphics[width=\linewidth]{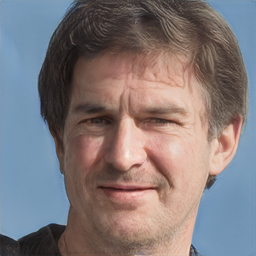}
     \includegraphics[width=\linewidth]{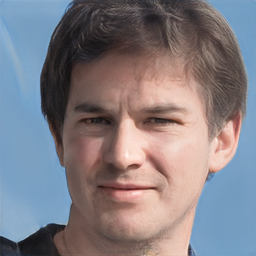}
     \end{minipage}
     }
   \hspace{-2.8mm}
   \subfloat[pSp]{
     \begin{minipage}{0.18\linewidth}
     \includegraphics[width=\linewidth]{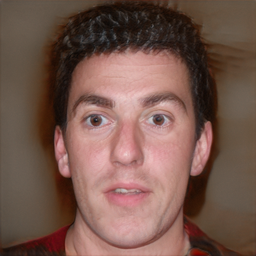}
     \includegraphics[width=\linewidth]{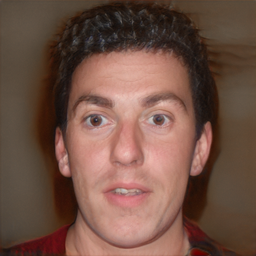}
     \includegraphics[width=\linewidth]{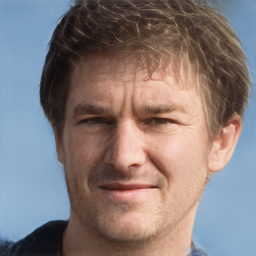}
     \includegraphics[width=\linewidth]{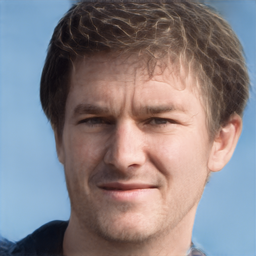}
     \end{minipage}
     }
   \hspace{-2.8mm}
   \subfloat[InD]{
     \begin{minipage}{0.18\linewidth}
     \includegraphics[width=\linewidth]{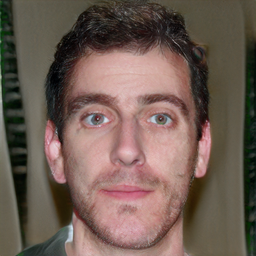}
     \includegraphics[width=\linewidth]{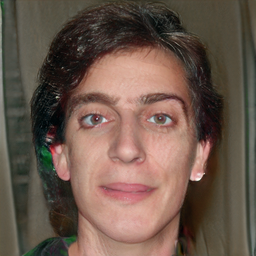}
     \includegraphics[width=\linewidth]{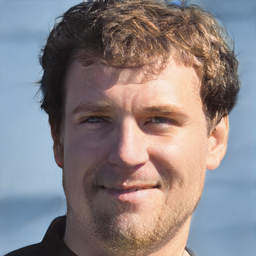}
     \includegraphics[width=\linewidth]{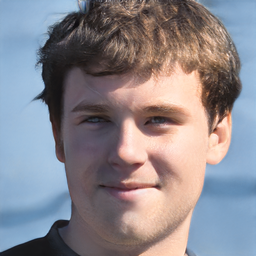}
     \end{minipage}
     }
   \hspace{-2.8mm}
    \subfloat[Ours]{
     \begin{minipage}{0.18\linewidth}
     \includegraphics[width=\linewidth]{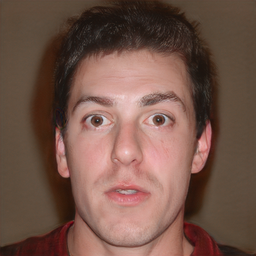}
     \includegraphics[width=\linewidth]{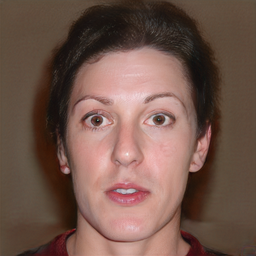}
     \includegraphics[width=\linewidth]{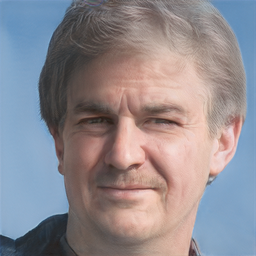}
     \includegraphics[width=\linewidth]{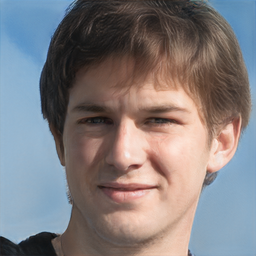}
     \end{minipage}
     }
    \rotatebox[origin=c]{90}{  --  \hspace{10mm} Age \hspace{10mm} +  \hspace{10mm} + \hspace{10mm} Gender \hspace{10mm} --}
\vspace{-3mm}\caption{Qualitative comparison on semantic edit on the nonlinear dataset.}
\label{fig_edit}\vspace{-6mm}
\end{figure*}

To further demonstrate our method is not restricted with the linear-based editing, we synthesize a new dataset under the nonlinear constraints by StyleFlow~\cite{abdal2021styleflow}. It consists of 1,000 sequences resulting in 5,000 images, with different semantic changes, such as pose, illumination, expression, eyeglasses, gender, and age. We conduct an inversion experiment on it and the results are shown in the Tab.~\ref{table}, Fig.~\ref{fig_rec}. Our method can obtain an accurate reconstruction on this nonlinear dataset. Besides, we also conduct nonlinear semantic editing task using the nonlinear StyleFlow~\cite{abdal2021styleflow}, as shown in the right parts of Tab.~\ref{table} and Fig.~\ref{fig_edit}. Our results are more similar with the GT (see the hair color of ``\emph{age+}'' in the right-bottom corner of Fig.~\ref{fig_edit}). These results prove that our method is not constrained by linear-editing assumption. That because our method is an \emph{optimization-based} GAN inversion method that does not rely on any attribute constraint for the input images, and the optimization is image-specific \emph{without training} a general network.

\begin{figure*}[t]
   \centering
    \subfloat[Original]{
    \begin{minipage}{0.14\linewidth}
     \includegraphics[width=\linewidth]{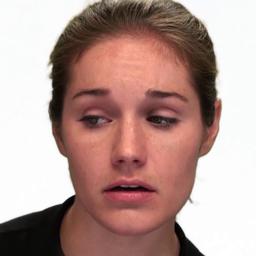}
     \includegraphics[width=\linewidth]{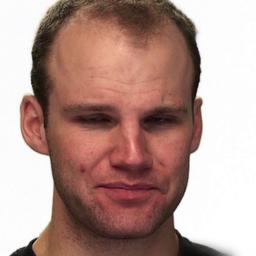}
     \includegraphics[width=\linewidth]{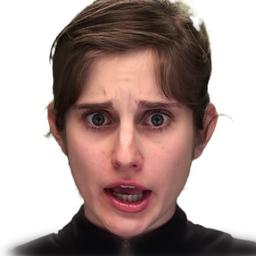}
     \includegraphics[width=\linewidth]{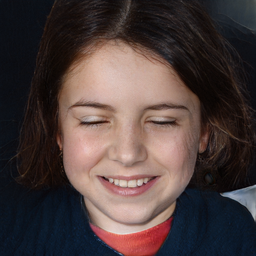}
     \includegraphics[width=\linewidth]{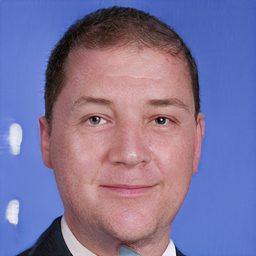}
     \includegraphics[width=\linewidth]{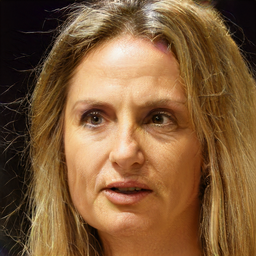}
     \end{minipage}
     }
   \hspace{-2.8mm}
   \subfloat[I2S]{
     \begin{minipage}{0.14\linewidth}
     \includegraphics[width=\linewidth]{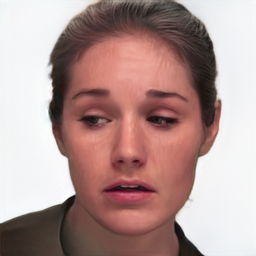}
     \includegraphics[width=\linewidth]{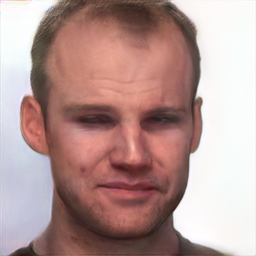}
     \includegraphics[width=\linewidth]{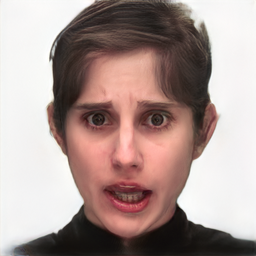}
     \includegraphics[width=\linewidth]{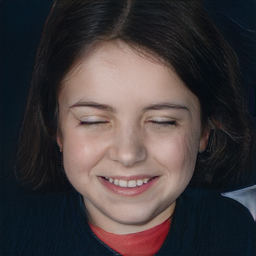}
     \includegraphics[width=\linewidth]{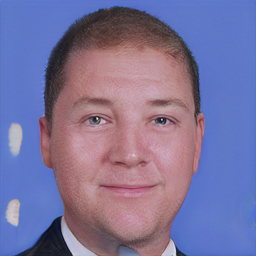} 
     \includegraphics[width=\linewidth]{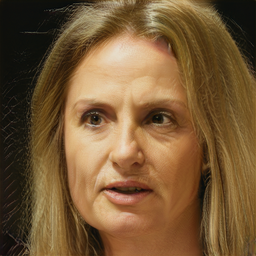}    
     \end{minipage}
     }
   \hspace{-2.8mm}
   \subfloat[pSp]{
     \begin{minipage}{0.14\linewidth}
     \includegraphics[width=\linewidth]{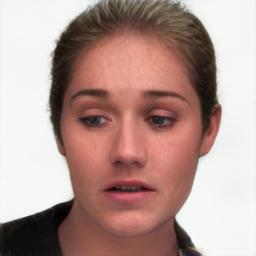}
     \includegraphics[width=\linewidth]{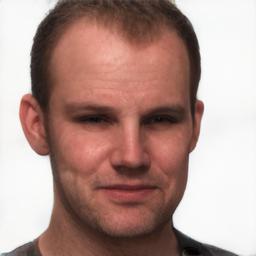}
     \includegraphics[width=\linewidth]{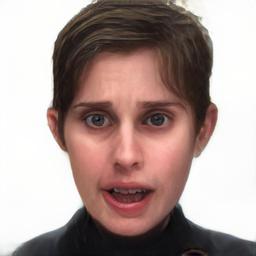}
     \includegraphics[width=\linewidth]{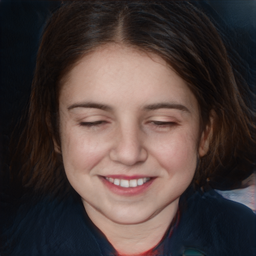}
     \includegraphics[width=\linewidth]{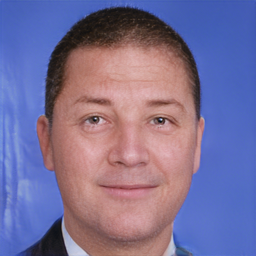}
     \includegraphics[width=\linewidth]{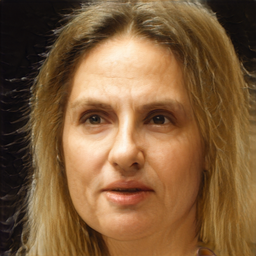}
     \end{minipage}
     }
   \hspace{-2.8mm}
   \subfloat[InD]{
     \begin{minipage}{0.14\linewidth}
     \includegraphics[width=\linewidth]{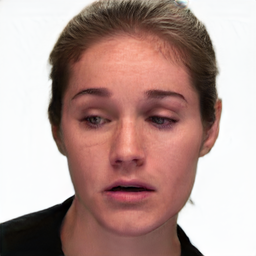}
     \includegraphics[width=\linewidth]{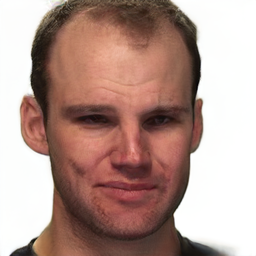}
     \includegraphics[width=\linewidth]{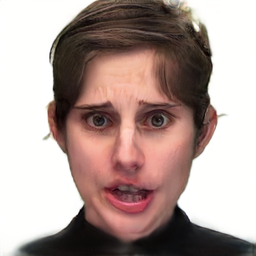}
     \includegraphics[width=\linewidth]{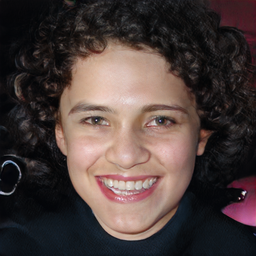}
     \includegraphics[width=\linewidth]{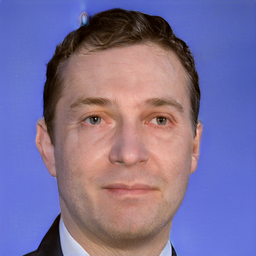}
     \includegraphics[width=\linewidth]{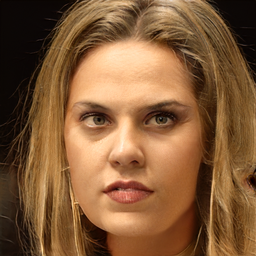}
     \end{minipage}
     }
   \hspace{-2.8mm}
    \subfloat[Ours]{
     \begin{minipage}{0.14\linewidth}
     \includegraphics[width=\linewidth]{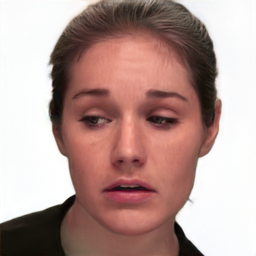}
     \includegraphics[width=\linewidth]{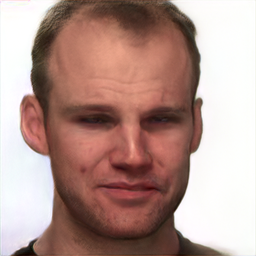}
     \includegraphics[width=\linewidth]{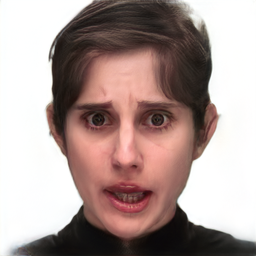}
     \includegraphics[width=\linewidth]{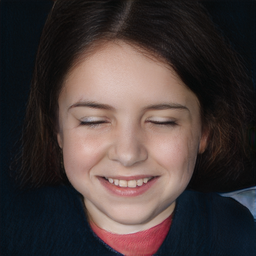}
     \includegraphics[width=\linewidth]{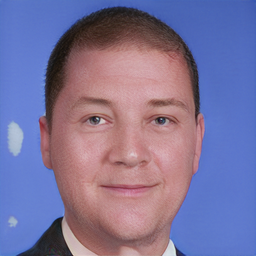}
     \includegraphics[width=\linewidth]{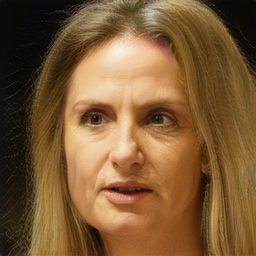}
     \end{minipage}
     }
   % \hspace{-.8mm}
   \hspace{-1.5mm}
   \rotatebox[origin=c]{90}{\rev{- - - - - - - - - - - - - - - - - - - - - - - - - - - - - - - - - - - - - - - - - - - - - - - - - - - - - - - - - - - - - - - - - - - - - - - - - - -}}
   \hspace{-3mm}
   \subfloat[I2S++]{
     \begin{minipage}{0.14\linewidth}
     \includegraphics[width=\linewidth]{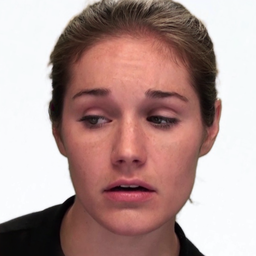}
     \includegraphics[width=\linewidth]{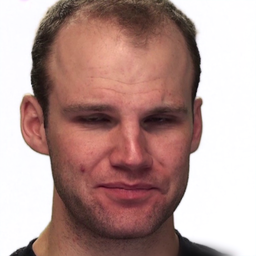}
     \includegraphics[width=\linewidth]{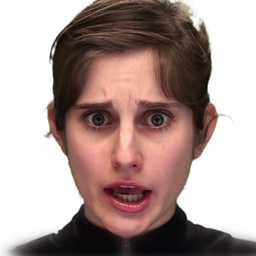}
     \includegraphics[width=\linewidth]{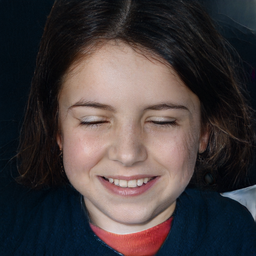}
     \includegraphics[width=\linewidth]{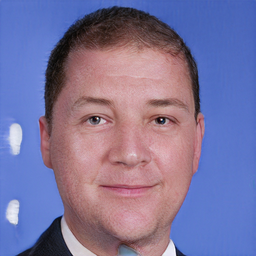}
     \includegraphics[width=\linewidth]{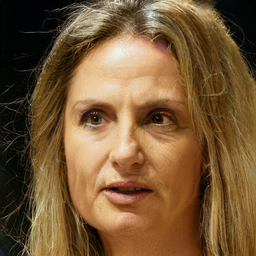}
     \end{minipage}
    }
   \hspace{-2.8mm}
   \subfloat[Ours++]{
     \begin{minipage}{0.14\linewidth}
     \includegraphics[width=\linewidth]{figure/inversion/truth/02-02-04-01-01-01-08_000124}
     \includegraphics[width=\linewidth]{figure/inversion/truth/02-02-06-01-01-02-19_000115}
     \includegraphics[width=\linewidth]{figure/inversion/truth/02-02-06-02-01-01-24_000066}
     \includegraphics[width=\linewidth]{figure/inversion/truth/000013_2}
     \includegraphics[width=\linewidth]{figure/inversion/truth/000010_2}
     \includegraphics[width=\linewidth]{figure/inversion/truth/000067_2}
     \end{minipage}
     }

\caption{Qualitative comparison on image reconstruction.}
\label{fig:reconstruction_exp}\vspace{-4mm}
\end{figure*}

We also give more qualitative comparison on image reconstitution on RAVDESS-12 Dataset and the linear-based Synthesized Dataset in Fig.~\ref{fig:reconstruction_exp}. We can see that our method can reconstruct the most faithful appearances by optimization latent code in the~$\mathcal{W+}$ space. Involving the~$\mathcal{N}$ space largely improves reconstruction quality and Ours++ can reconstruct the correct colors.

\begin{figure*}[t]
   \centering
    \hspace{-5mm}
    % \rotatebox[origin=c]{90}{  \textbf{+}  \hspace{15mm} \textbf{Age}   \hspace{15mm} \textbf{--} \hspace{15mm} \textbf{+}  \hspace{15mm} \textbf{EyeGlasses} \hspace{15mm}  \textbf{--} \hspace{20mm} \textbf{+}  \hspace{20mm} \textbf{Gender} \hspace{10mm}  \textbf{--}}
    \rotatebox[origin=c]{90}{  \textbf{+}  \hspace{15mm} \textbf{Age}   \hspace{15mm} \textbf{--} \hspace{15mm}  \textbf{+}  \hspace{20mm} \textbf{Gender} \hspace{10mm}  \textbf{--}}
    \subfloat[GT]{
    \begin{minipage}{0.195\linewidth}
     \includegraphics[width=\linewidth]{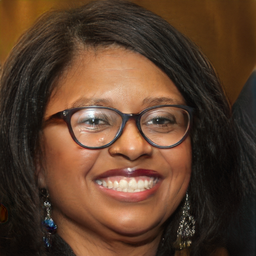}
     \includegraphics[width=\linewidth]{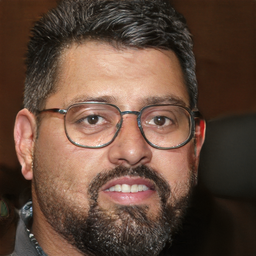}
     \includegraphics[width=\linewidth]{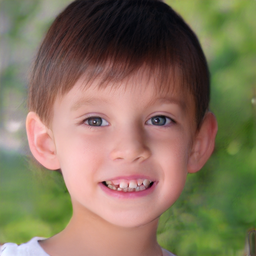}
     \includegraphics[width=\linewidth]{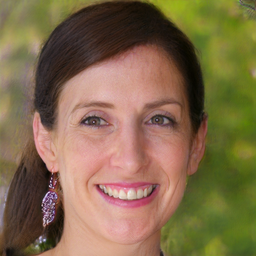}
     \end{minipage}
     }
   \hspace{-2.8mm}
   \subfloat[I2S]{
     \begin{minipage}{0.195\linewidth}
     \includegraphics[width=\linewidth]{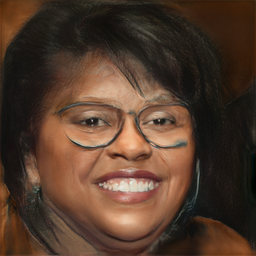}
     \includegraphics[width=\linewidth]{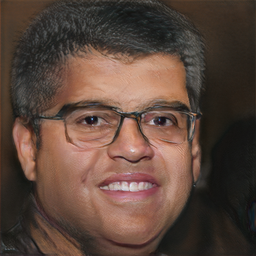}
     \includegraphics[width=\linewidth]{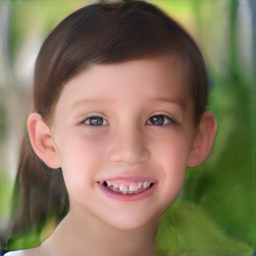}
     \includegraphics[width=\linewidth]{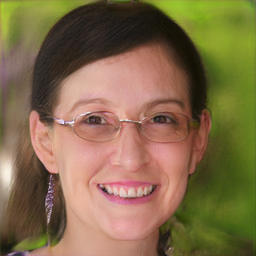}

     \end{minipage}
     }
   \hspace{-2.8mm}
    \subfloat[Ind]{
     \begin{minipage}{0.195\linewidth}
     \includegraphics[width=\linewidth]{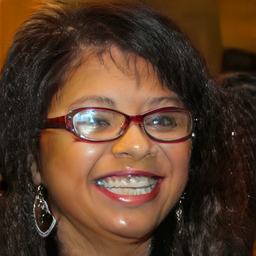}
     \includegraphics[width=\linewidth]{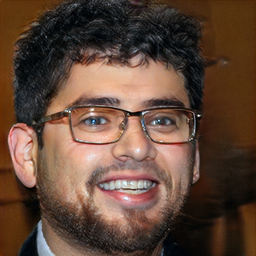}
     \includegraphics[width=\linewidth]{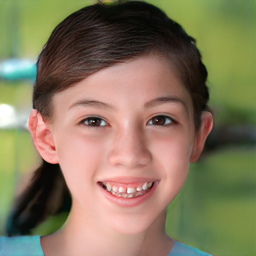}
     \includegraphics[width=\linewidth]{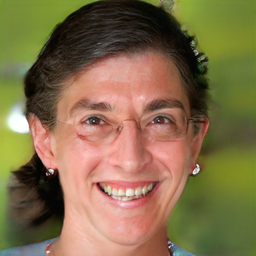}

     \end{minipage}
     }
   \hspace{-2.8mm}
    \subfloat[pSp]{
     \begin{minipage}{0.195\linewidth}
     \includegraphics[width=\linewidth]{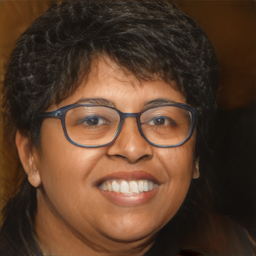}
     \includegraphics[width=\linewidth]{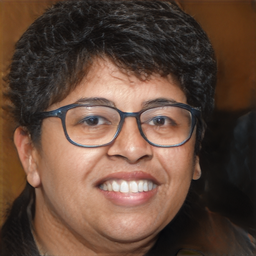}
     \includegraphics[width=\linewidth]{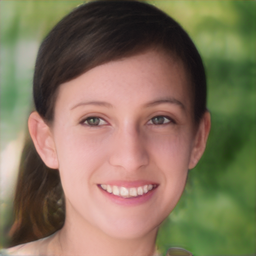}
     \includegraphics[width=\linewidth]{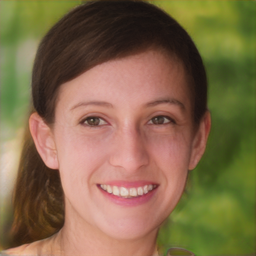}

     \end{minipage}
     }
   \hspace{-2.8mm}
   \subfloat[Ours]{
     \begin{minipage}{0.195\linewidth}
     \includegraphics[width=\linewidth]{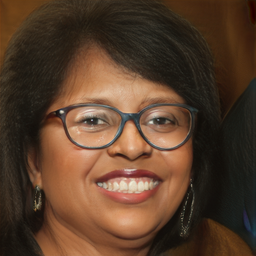}
     \includegraphics[width=\linewidth]{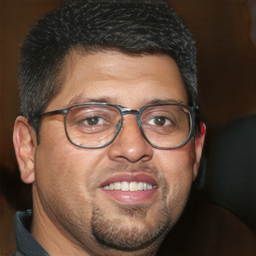}
     \includegraphics[width=\linewidth]{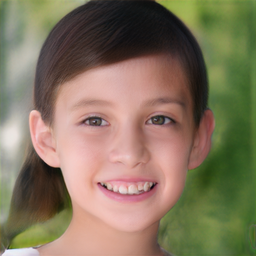}
     \includegraphics[width=\linewidth]{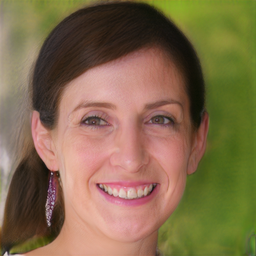}

     \end{minipage}
     }
   \hspace{-2.8mm}
\vspace{-3mm}\caption{Qualitative comparison on semantic editing.}
\label{fig:semantic_exp_g}\vspace{-3mm}
\end{figure*}

The qualitative comparison on image editing task on Synthesized dataset can be seen in Fig.~\ref{fig:semantic_exp_g}. Our edited results are more similar with the ground truths and show the pleasure appearances, which indicates that our inverted latent codes are close enough with the GT codes.

% {\small
% \bibliographystyle{ieee_fullname}
% \bibliography{egbib2}
% }